\documentclass[12pt]{iopart}

\usepackage{amsmath}
\usepackage{amsfonts}

\usepackage{float}
\usepackage{graphicx}
\usepackage{caption}
\usepackage{subcaption}

\usepackage{hyperref}       
\hypersetup{colorlinks,linkcolor=blue, citecolor=red}
\usepackage{xcolor}

\newcommand{\unit}{1\!\!1}

\def\<{\langle} \def\>{\rangle}

\DeclareMathOperator*{\argmin}{arg\,min}

\usepackage{amsthm}

\makeatletter
\newcommand\appendix@section[1]{%
  \refstepcounter{section}%
  \orig@section*{Appendix \@Alph\c@section: #1}%
  \addcontentsline{toc}{section}{\@Alph\c@section: #1}%
}
\let\orig@section\section
\g@addto@macro\appendix{\let\section\appendix@section}
\makeatother

\begin{document}

\title[Probing transfer learning with a model of synthetic correlated datasets]{Probing transfer learning with a model of synthetic correlated datasets}

%

\author{
Federica Gerace$^\dag$, Luca Saglietti$^\dag$, Stefano Sarao Mannelli$^{\ddag}$,\\
Andrew Saxe$^{\ddag,\S}$, and Lenka Zdeborov\'a$^\dag$
}
\address{
$\dag$ \textit{SPOC laboratory, \'Ecole Polytechnique Fédérale de Lausanne (EPFL)
}\\
$\ddag$ \textit{Gatsby Computational Neuroscience Unit \& Sainsbury Wellcome Centre, University College London}\\
$\S$ \textit{CIFAR Azrieli Global Scholars programme, CIFAR}
}

\begin{abstract}
    Transfer learning can significantly improve the sample efficiency of neural networks, by exploiting the relatedness between a data-scarce target task and a data-abundant source task. 
    Despite years of successful applications, transfer learning practice often relies on ad-hoc solutions, while theoretical understanding of these procedures is still limited. In the present work, we re-think a solvable model of synthetic data as a framework for modeling correlation between data-sets. This setup allows for an analytic characterization of the generalization performance obtained when transferring the learned feature map from the source to the target task. Focusing on the problem of training two-layer networks in a binary classification setting, we show that our model can capture a range of salient features of transfer learning with real data. Moreover, by exploiting parametric control over the correlation between the two data-sets, we systematically investigate under which conditions the transfer of features is beneficial for generalization.
\end{abstract}

%
%
%
%
%

Paper published in \textit{Machine Learning: Science and Technology}, please cite as: \\
``{Gerace et al 2022 Mach. Learn.: Sci. Technol. \href{https://doi.org/10.1088/2632-2153/ac4f3f}{https://doi.org/10.1088/2632-2153/ac4f3f}}" .

\setcounter{secnumdepth}{2}
\setcounter{tocdepth}{2}
\tableofcontents{}

\title[Probing transfer learning with a model of synthetic correlated datasets]{Probing transfer learning with a model of synthetic correlated datasets}

\section{Introduction}


The incredible successes of deep learning originate from years of developments on network architectures and learning algorithms, but also from increasing amounts of data available for training. In many deep learning applications, data abundance allows for an extensive empirically-driven exploration of models and hyper-parameter tuning to fit billions of model parameters \cite{krizhevsky2012imagenet,he2016deep,brown2020language}.
Despite efforts in building data-efficient pipe-lines \cite{han2015deep,tan2019efficientnet}, deep networks remain intrinsically data-hungry \cite{chen2014big}.
In some settings, however, this procedure is impractical, due to the limited size of the dataset. For instance, the bureaucracy behind the acquisition of medical data and the lack of automated systems for their labelling complicate the deployment of deep learning models in medical imaging \cite{frid2018gan,shin2018medical,zhao2019data}.

A solution that can drastically reduce the need for new labelled data is \emph{transfer learning} \cite{thrun2012learning, shin2016deep,raghu2019transfusion}. This technique aims to improve the generalization performance of a network trained on a data-scarce \emph{target} task, by leveraging the information that a second network has acquired on a related and data-abundant \emph{source} task. The idea is that the network does not need to learn to recognize the relevant features of data from scratch, but can start from an advanced, yet imperfect, stage of learning. In the simplest formulation of transfer learning, the first layers of a neural network--those that perform feature extraction--are transferred and kept fixed, and only the last segment of the network is retrained on the new dataset. In practice, an additional end-to-end fine-tuning stage is often added \cite{chollet2020}, to adapt the learned features to the target task.

Despite its great success and its extensive use in deep learning applications, theoretical understanding of transfer learning is still limited and many practical questions remain open. For instance: \textit{How related do the source and target tasks need to be? Is it better to transfer from complex to simpler datasets or vice versa? How much data is necessary in the source dataset to make the transfer effective? How does the size of the transferred feature map impact performance? And when is the fine-tuning stage beneficial?}  

A key ingredient for gaining theoretical insight on these questions is to obtain a generative model for structured data able to produce non-trivial correlations between different tasks. The influence of the structure of data on learning has been an active research topic in recent years \cite{goldt2019modelling, gerace2020generalisation, pastore2020statistical, d2021more}. An important step forward was taken in \cite{goldt2019modelling}, where it was shown that a simple generative model called the \textit{hidden manifold model} (HMM) induces learning behaviors that more closely resemble those observed on real data. The HMM is built on the idea that real-world datasets are intrinsically low dimensional, as recent studies have shown for some of the most widely used benchmarks. For instance, the intrinsic dimensions \cite{facco2017estimating} of MNIST and CIFAR10 were estimated to be $L\approx15$ and $L\approx35$ \cite{spigler2019asymptotic}, compared to the actual number of pixels $D = 784$ and $D = 1024$. This concept is also demonstrated in modern generative models \cite{kingma2013auto, goodfellow2014generative}, where high-dimensional data is produced starting from lower-dimensional random vectors.

\paragraph*{Contributions:} In the present work, we propose a framework for better understanding transfer learning, where the many variables at play can be isolated in controlled experiments. 
\begin{itemize}
    \item In Sec.~\ref{sec:problem}, we introduce the correlated hidden manifold model (CHMM), a tractable setting where the correlation between datasets becomes explicit and tunable. 
    \item By employing methods developed in \cite{loureiro2021capturing}, in Sec.~\ref{sec:theory}, we analytically characterize the asymptotic generalization error of transfer learning (without fine-tuning) on the CHMM, where learned features are transferred from source to target task and are kept fixed while training on the second task.
    \item We demonstrate similar behavior of transfer learning in the CHMM and in experiments on real data, in Figs.~\ref{fig:teacher_pert_real_vs_synthetic} and \ref{fig:asymmetric_real_vs_synthetic}. 
    \item In Sec.~\ref{sec:result} we compare the performance of transfer learning with the generalization error obtained by three alternative learning models: a random feature model; a network trained from scratch on the target task; and transfer learning with an additional stage of fine-tuning. 
    \item In Figs.~\ref{fig:teacher_pertubations_phase_diagram} and \ref{fig:asymmetric_phase_diagram} we leverage parametric control over the correlation between source and target datasets to extensively explore different transfer learning scenarios and delineate the boundaries of effectiveness of the feature transfer. We also trace possible \textit{negative transfer} effects observed on real data, such as over-specialization and overfitting \cite{yosinski2014transferable, kornblith2019better} in Fig.~\ref{fig:teacher_pertubations_phase_diagram}.
\end{itemize}

\paragraph*{Further related work:} 

Starting from the seminal work of \cite{baxter2000model}, the statistical learning community has produced a number of theoretical results bounding the performance of transfer learning, often approaching the problem from closely related settings like multi-task learning, few-shot learning, domain adaptation and hypothesis transfer learning (see \cite{zhang2019transfer,wang2020generalizing,zhang2021survey} for surveys). Most of these results rely on standard proof techniques based on the Vapnik-Chervonenkis dimension, covering number, stability, and Rademacher complexity. A recent notable example of this type of approach can be found in \cite{du2020few}. Another recent work considers a two-stage setting similar to the one analyzed in the present paper, where the feature map learned on a first source task is transferred but not fine-tuned on a target task \cite{tripuraneni2020theory}. However, these works focus on worst-case analyses, whereas our work has a complementary focus on characterizing average-case behavior, i.e., typical transfer learning performance that we believe may be closer to observations in practice. 

An alternative -- yet less explored -- approach is to accept stronger modeling assumptions in order to obtain descriptions of typical learning scenarios. Three recent works rely on the teacher-student paradigm \cite{engel2001statistical}, 
exploring transfer learning in deep linear networks \cite{lampinen2018analytic} and in single-layer networks \cite{dar2020double,dhifallah2021phase}. Despite many differences, these works confirm the intuitive finding that fewer examples are needed to achieve good generalization when the teachers of source and target tasks are more aligned. While these studies provide a careful characterization of the effect of initialization inherited from the source task, they do not address other crucial aspects of transfer learning, especially the role played by feature extraction and feature sharing among different tasks. In modern applications feature reuse is key in the transfer learning boost \cite{raghu2019rapid} and it is lacking in the current modeling effort.
This is precisely what we investigate in the present work.

\section{Definition of the correlated hidden manifold model}\label{sec:problem}

\begin{figure}[]
    \centering
    \includegraphics[width=\linewidth]{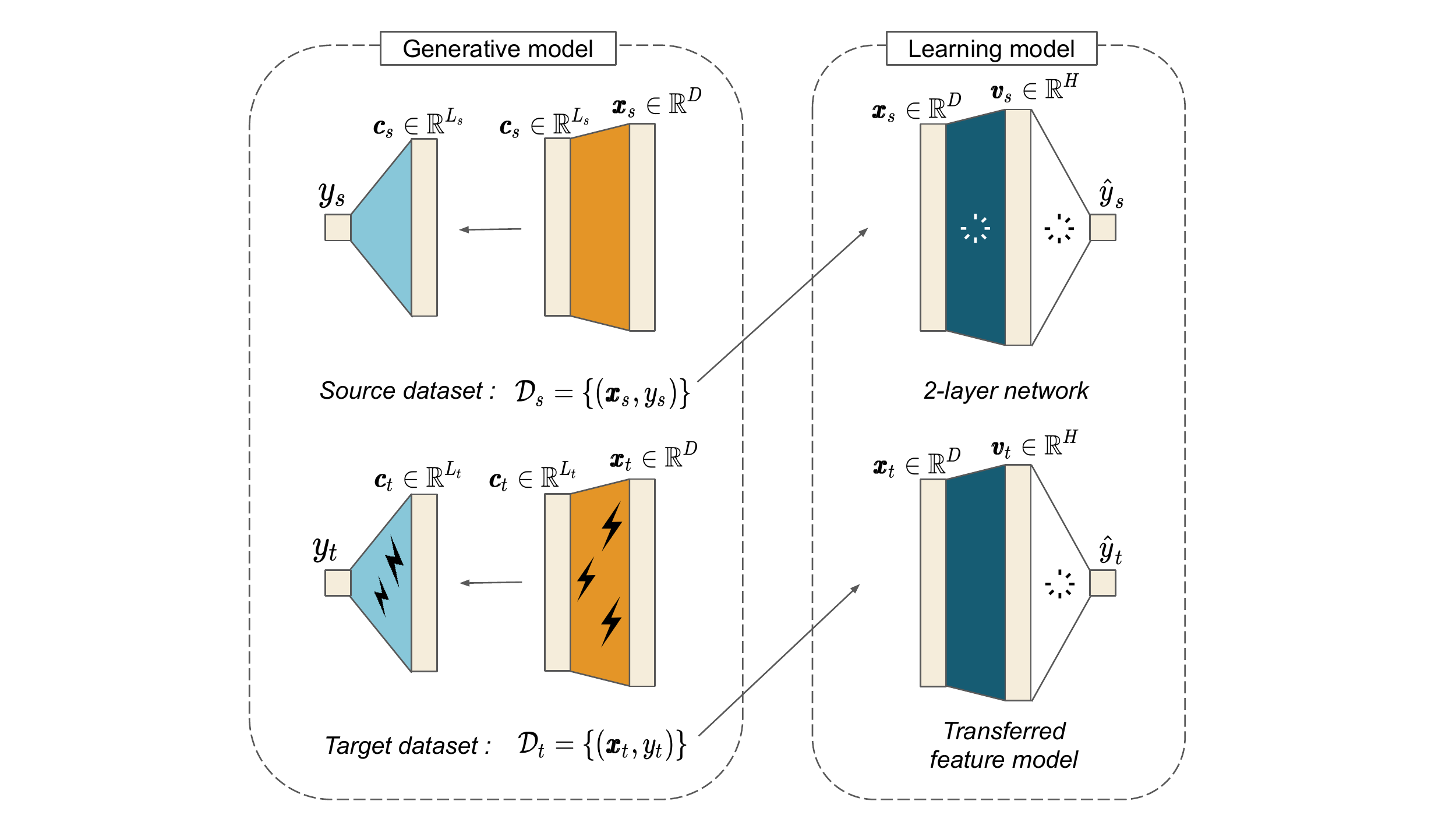}
    \caption{\textbf{Diagram of the CHMM.} Left panel: the generative model consists of two hidden manifold models that share features (orange) and teacher weights (light blue), with some perturbations (lightning icons). Right panel: a two-layer network is trained on the source dataset, and the learned feature map (dark blue) is transferred to the second architecture -- the transferred feature model. The loading symbols indicate which layer is trained in each phase.
    }
    \label{fig:model_cartoon}
\end{figure}

We propose the correlated hidden manifold model (CHMM) as a high-dimensional tractable model of transfer learning, sketched in Fig.~\ref{fig:model_cartoon}. 
To capture the phenomenology of transfer learning, two key ingredients are necessary. First, we need to express the relationship between source and target tasks, by defining a \emph{generative model} for correlated and structured datasets. Second, we need to specify the \emph{learning model} and the associated protocol through which the feature map is transferred over the two tasks. In the following, we describe both parts in detail.  

\subsection{Generative model.} The relationship between classification tasks can take different forms. Two tasks could rely on different features of the input, but these features could be read-out identically to produce the label; or two tasks could use the same features, but require these to be read-out according to different labeling rules. In the \emph{hidden manifold model} (HMM) \cite{goldt2019modelling}, explicit access to a set of generative features and to the classification rule producing the data allows us to model and combine correlations of both kinds.

The correlated-HMM combines two HMMs -- representing source and target datasets -- in order to introduce correlations at the level of the generative-features and the labels. The key element in the modeling is thus the relation between the two datasets.
On a formal level, we construct source and target datasets, $\mathcal{D}_s = \{ \boldsymbol{x}^{\mu}_s, y^{\mu}_s \}_{\mu = 1}^{M_s}$ and $\mathcal{D}_{t} = \{ \boldsymbol{x}^{\mu}_{t}, y^{\mu}_{t} \}_{\mu = 1}^{M_{t}}$, as follows: let $D$ denote the input dimension of the problem (e.g., the number of pixels), and $L_s$ the latent (intrinsic) dimension of the source dataset. First, we generate a pair $(\boldsymbol{F}_s, \boldsymbol{\theta}_s)$, where $\boldsymbol{F}_s\in \mathbb{R}^{L_s\times D}$ is a matrix of $D$-dimensional Gaussian generative features, $(\boldsymbol{F}_s)_{ij}\sim\mathcal{N}(0,1)$, and $\boldsymbol{\theta}_s \in \mathbb{R}^{L_s}$ denotes the parameters of a single-layer teacher network, $(\boldsymbol{\theta}_s)_i \sim \mathcal{N}(0,1)$.
The input data points for the source task, $\boldsymbol{x}^{\mu}_s \in \mathbb{R}^{D}$, are then obtained as a combination of the generative features with Gaussian coefficients $\boldsymbol{c}^\mu_s \in \mathbb{R}^{L_s}$, $(\boldsymbol{c}^\mu_s)_i \sim \mathcal{N}(0,1)$, while the binary labels $y^\mu\in\{-1,1\}$ are obtained as the output of the teacher network, acting directly on the coefficients:
\begin{equation} \label{eq:HMM}
    \boldsymbol{x}^{\mu}_s = \mbox{ReLU} \left(\frac{\boldsymbol{c}^{\mu}_s \boldsymbol{F}_{{s}}}{\sqrt{L_s}} \right); \hspace{20mm}
    y_s^{\mu} = \mbox{sign} \left(\frac{\boldsymbol{c}^{\mu}_s \boldsymbol{\theta}_{{s}}}{\sqrt{L_s}} \right).
\end{equation}
Thus, starting from a typically lower-dimensional latent vector $\boldsymbol{c}^\mu_s$ the model produces structured inputs embedded in a high-dimensional space, as in many modern generative models \cite{kingma2013auto, goodfellow2014generative}. Note that the choice of the $\mbox{ReLU}$ non-linearity is not a necessary requirement of the proposed modeling framework (which would allow any choice of common activation functions), and that sparse inputs are found in many standard benchmarks (e.g. MNIST-like datasets \cite{deng2012mnist, cohen2017emnist, notmnist}). On the other hand, the Gaussianity of the generative features and coefficients, and of the teacher vector, is a customary but strong modeling assumption which is crucial for the theoretical treatment described in the following sections. 

In order to construct the pair $(\boldsymbol{F}_t, \boldsymbol{\theta}_t)$ for the target task, we directly manipulate both the features and the teacher network of the source task. To this end, we consider three families of transformations:
\begin{itemize}
    \item \emph{Feature perturbation and substitution}. This type of transformation can be used to model correlated datasets that may differ in style, geometric structure, etc. In the CHMM, we regulate these correlations with two independent parameters: the parameter $\eta$, measuring the amount of noise injected in each feature, and the parameter $\rho$, representing the fraction of substituted features:
    \begin{equation}
    (\boldsymbol{F}_t)_{ij} = 
    \begin{cases}
        & ({\boldsymbol{\tilde{F}}})_{ij} \quad\quad\quad\quad\quad\quad\quad\quad\,\,\quad\hspace{2mm} \text{for  $i=1,\dots,\rho L_s$}
        \\
        & \eta (\boldsymbol{F}_s)_{ij} + \sqrt{1 - \eta^2} (\boldsymbol{\tilde{F}})_{ij}  \quad\quad \text{for  $i=\rho L_s+1,\dots,L_s$}
    \end{cases}, ~~~(\boldsymbol{\tilde{F}})_{ij} \sim \mathcal{N}(0,1);
    \end{equation}
    \item \emph{Teacher network perturbation}. This type of transformation can be used to model datasets with the same set of inputs but grouped into different categories. In the CHMM, we represent this kind of task misalignment through a perturbation of the teacher network with parameter~$q$:
    \begin{equation}
      (\boldsymbol{\theta}_{{t}})_i = q (\boldsymbol{\theta}_{{s}})_i + \sqrt{1 - q^2} ({\boldsymbol{\tilde{\theta}}})_i\,\,, \quad\quad\quad ({\boldsymbol{\tilde{\theta}}})_i \sim \mathcal{N}(0,1) ;
    \end{equation}
    \item \emph{Feature addition or deletion}. This type of transformation can be used to model datasets with different degrees of complexity and thus different intrinsic dimensions. In the CHMM, we alter the latent space dimension  $L_t\neq L_s$, adding or subtracting some features from the generative model:
    \begin{equation}
    (\boldsymbol{F}_t)_{ij} = 
    \begin{cases}
        & (\boldsymbol{F}_{s})_{ij} \hspace{4mm} \text{for  $i = 1,\dots,\min(L_s,L_{t})$}
        \\
        & (\boldsymbol{\tilde{F}})_{ij} \hspace{5mm} \text{for  $i=\min(L_s,L_{t})+1,\dots,L_{t}$}
    \end{cases}, ~~~(\boldsymbol{\tilde{F}})_{ij} \sim \mathcal{N}(0, 1).
    \end{equation}
    Moreover, the target teacher vector will also have a different number of components: 
    \begin{equation}
    (\boldsymbol{\theta}_{t})_i = 
    \begin{cases}
        & (\boldsymbol{\theta}_{s})_{i} \hspace{4mm} \text{for  $i = 1,\dots,\min(L_s,L_{t})$}
        \\
        & (\boldsymbol{\tilde{\theta}})_i \hspace{5.5mm} \text{for  $i=\min(L_s,L_{t})+1,\dots,L_{t}$}
    \end{cases}, ~~~(\boldsymbol{\tilde{\theta}})_i \sim \mathcal{N}(0, 1).
    \end{equation}
\end{itemize}
The target dataset is again produced according to eq.\,\ref{eq:HMM}, with the pair $(\boldsymbol{F}_t, \boldsymbol{\theta}_t)$ and starting from i.i.d. Gaussian latent vectors $\boldsymbol{c}^\mu_t \in \mathbb{R}^{L_t}$, $(\boldsymbol{c}^\mu_t)_i\sim\mathcal{N}(0,1)$.  

We can easily identify counterparts of these types of transformations in real world data. For example, a simple instance of datasets that slightly differ in style but share the same labeling (as modeled through feature perturbation/substitution in the CHMM) can be obtained by applying a data-augmentation transformation like elastic distortions on a given dataset. 
Another typical transfer learning setting is one where the same data is clustered according to different labeling rules (teacher perturbation in the CHMM), e.g. \emph{even-odd digits} vs \emph{digits greater-smaller than 5} in the MNIST dataset. 
Finally, one could obtain two datasets that only differ in the complexity level (and thus in the latent space dimension) by selecting a slice of a given dataset that only contains a subset of the original categories. In this way some of the recognizable traits of the images in the full dataset will no longer appear in the reduced one (as in the feature deletion case in the CHMM). In section~\ref{sec:result}, we provide evidence of the qualitative agreement between the phenomenology observed in the CHMM of task correlation and in real-world datasets.  

\subsection{Learning model.}
We specialize our analysis to the case of two-layer neural networks. This is the simplest architecture able to perform feature extraction and develop non-trivial representations of the inputs through learning. Note that the learning problem created through the HMM construction could also be tackled with a single layer network \cite{gerace2020generalisation}, but the absence of a transferable feature map would not allow for the present study. We will denote with $H$ the number of hidden units in the network and focus on the case of $\mathrm{ReLU}$ activation functions (again, this choice is purely arbitrary). 
To train the network, we employ a standard binary cross-entropy loss with $L_2$-regularization on the second layer.

We define the transfer protocol as follows. First, we randomly initialize a two-layer network with i.i.d.$\sim\mathcal{N}(0, 1)$ weights and train it on the source classification task (following \cite{kornblith2019better}, early stopping is employed in this first training step). Then, we transfer the learned feature map, i.e. the first-layer weights, $\boldsymbol{\hat{w}}_1 \in \mathbb{R}^{H\times D}$, to a second two-layer network. Finally, we only train the last layer of the second network, $\boldsymbol{w}_2 \in \mathbb{R}^H$, while keeping the first layer frozen, on the target task with loss:
\begin{equation}
	{\boldsymbol{\hat w}_2} = \underset{\boldsymbol{w}_2}{\argmin}~\left[ 
  \sum\limits_{\mu=1}^{M_t}\ell\left(y^{\mu}, \frac{\boldsymbol{v}^\mu_t\cdot
  \boldsymbol{w}_2}{\sqrt{H}}\right) + \frac{\lambda}{2}||\boldsymbol{w}_2||_{2}^2 \right],
  \hspace{10mm} 
  \boldsymbol{v}^\mu_t = \text{ReLU} \left( \frac{\boldsymbol{x}^\mu_t\cdot
  \boldsymbol{\hat{w}}_1}{D} \right),
\end{equation}
where $\ell(y,x) = \log \left(1 + \exp \left( - yx\right) \right)$ is the binary cross-entropy. We refer to this learning pipeline as the \textit{transferred feature model} (TF). This notation puts the TF in contrast to the well-studied \textit{random feature model} (RF) \cite{rahimi2007random,mei2019generalization}, where the first-layer weights $\boldsymbol{w}_{1}$ are again fixed but random, i.e. sampled i.i.d.$\sim\mathcal{N}(0, 1)$. Of course, when the correlation between source and target tasks is sufficient, TF is expected to outperform RF.

Throughout this work, we will generally assume the hidden-layer dimension $H$ to be of the same order of the input size $D$. Note that, when the width of the hidden layer diverges, different limiting behaviors could be observed \cite{geiger2020disentangling}.

\section{Theoretical framework}
\label{sec:theory}

One of the main points of the CHMM is that the typical generalization performance associated with the above described transfer learning protocol can be characterized analytically in high-dimensions by means of the replica approach. This tool is a non-rigorous analytical method from statistical physics \cite{mezard1987spin}, used to characterize high-dimensional optimization problems through a narrow set of scalar fixed-point equations. In this reduced asymptotic description, the original dependence on the microscopic details of the model is captured through a set of overlap parameters, which are assumed to concentrate around their typical values in high dimensions. Note that, although the replica approach is heuristic in general settings, there is overwhelming numerical evidence of the exactness of the yielded results, and a rigorous proof has been derived in several models.  

A crucial aspect of the transfer learning setting analyzed in this paper, allowing the replica analysis described below, is that the TF model only learns the second layer weights $\boldsymbol{w}_2$ on the target task, while the feature map is kept fixed through training. In fact, it has been recently conjectured and empirically demonstrated in \cite{loureiro2021capturing} that there exists a broad universality class, featuring diverse learning models with fixed feature maps that asymptotically behave as simpler Gaussian covariate models, and that can be successfully studied by means of the replica approach. According to the generalized Gaussian equivalence theorem (GET) \cite{goldt2020gaussian}, in the high-dimensional limit $L,D,H,M\to\infty$ (defined in the previous paragraphs), with $\gamma=L/H$ and $\alpha = M/H$ of $\mathcal{O}(1)$, the generalization error of the two-layer network can be expressed as an easy-to-evaluate two-dimensional integral:
\begin{equation}
\lim_{D \to \infty} \epsilon_g = \mathbb{E}_{\delta, \nu} \left[\left( \mbox{sign}\left( \nu \right) -  \mbox{sign}\left( \delta \right)\right)^2\right]\label{eq:gen_err},
\end{equation}
where $\left(\delta, \nu\right)$ are jointly Gaussian random variables, with zero mean and covariance matrix
\begin{equation}
    \begin{bmatrix} \nu \\ \delta \end{bmatrix} \in \mathbb{R}^{2} \sim \mathcal{N}\left(0, \begin{bmatrix} \rho & m_{\star} \\ m_{\star} & q_{\star} \end{bmatrix} \right).
\end{equation}
The parameters $\rho = \left\lVert \boldsymbol{\theta}_t \right\rVert^2_2/D$ and $\left( m_{\star}, q_{\star} \right)$ are extremizers of the quenched free-entropy potential:
\begin{equation}
    \phi = \underset{q,V,m, \hat{q}, \hat{V}, \hat{m}}{\mathrm{extr}} \left\{\frac{1}{2}\left( q \hat{V} - \hat{q} V \right) - \sqrt{\gamma} m \hat{m} + \alpha g_E \left( q,V,m \right)+ g_S \left( \hat{q}, \hat{V}, \hat{m} \right)\right\}, \label{eq:free_entropy}
\end{equation}
where $q,V,m$ and $\hat{q},\hat{V},\hat{m}$ are the overlap parameters and their conjugates respectively.  The two functions $g_S$ and $g_E$ in Eq.~\eqref{eq:free_entropy} can be interpreted as competing entropic and energetic contributions. The entropic potential, 
\begin{equation}
    \label{eq:entropic_potential}
    g_S = \lim_{H \rightarrow \infty}\frac{1}{2H} \tr \left( \hat{m}^2 \Phi^T \boldsymbol{\theta}_t \boldsymbol{\theta}_t^T\Phi + \hat{q} \Omega \right) \left( \lambda \unit^{D\times D} + \hat{V} \Omega \right)^{-1},
\end{equation}
depends on the spectral properties of two matrices, $\Omega \in \mathbb{R}^{D \times D}$ and $\Phi \in \mathbb{R}^{L \times D}$, that encode -- through the generative coefficients $\boldsymbol{c}$ and the activations $\boldsymbol{v}$ (below) -- the non-trivial dependency on the correlations between training samples and the non-linear feature maps: 
\begin{equation}
    \begin{bmatrix} \boldsymbol{c} \\ \boldsymbol{v} \end{bmatrix} \in \mathbb{R}^{L+D} \sim \mathcal{N}\left(0, \begin{bmatrix} \unit^{L\times L} & \Phi \\ \Phi^T & \Omega \end{bmatrix} \right).\label{eq:c_and_v}
\end{equation}
From Eq.~\eqref{eq:entropic_potential} and \eqref{eq:c_and_v}  we can notice that, applying the GET, the non-linear learning problem with a fixed but generic feature map is mapped onto a simple Gaussian covariate model. 

The energetic potential $g_E$ is instead defined as
\begin{align}
    & g_E = \int \mathcal{D}z \, \mathcal{H}\left(-\frac{m \, z}{\sqrt{\rho q - m^2}}\right) \frac{1}{2}( M_E(1) +  M_E(-1) ),
    \\
    & M_E(y) = \max_u \left[-\frac{u^2}{2} -\ell\left(y, \sqrt{V} u + \sqrt{q}\right) \right];
\end{align}
with $\mathcal{H}(x) = \int_{-\infty}^\infty dz \frac{e^{-z^2/2}}{\sqrt{2\pi}}$ and $\ell(y,x)=\log(1+\exp(-y x))$ the binary cross-entropy loss. 

The extremizers of eq.~\eqref{eq:free_entropy} have a precise physical meaning, since they correspond to the typical values of practically measurable overlaps:
\begin{equation}
    q = \frac{1}{D} \boldsymbol{\hat{w}}_2^T \Omega \boldsymbol{\hat{w}}_2; \,\,\,\,\,\,\,\,\, 
    m = \frac{1}{\sqrt{LD}} \boldsymbol{\theta}_t^T \Phi \boldsymbol{\hat{w}}_2
\end{equation}
over different possible realizations of the training set.
Given the values of $m_{\star}$ and $q_{\star}$, one can finally compute the two-dimensional Gaussian integral in equation \eqref{eq:gen_err}, yielding the generalization error achieved at the end of training in closed-form expression:
\begin{equation}
    \epsilon_g = \frac{1}{\pi} \arccos \left( \frac{m_{\star}}{\sqrt{\rho \, q_{\star}}} \right).
\end{equation}

An important consideration is that these results are obtained in the so-called replica symmetric ansatz. Despite being the simplest possible ansatz for these kind of calculations, it is also known to be the correct one for convex problems like the logistic regression setting under study. We also remark that the equations above are already in the zero-temperature limit, which is the relevant one for studying optimization problems. This limit is non-trivial and requires the introduction of appropriate scaling laws. We refer the reader to \cite{loureiro2021capturing} for further details on the derivation.

In this work, we employ this set of equations to predict the generalization performance of the TF model, described in section~\ref{sec:theory}. We accept the validity of the Gaussian equivalence as a working assumption, setting up to verify its viability \emph{a posteriori} through comparison with numerical simulations. To analyze the transfer learning setting proposed in Section~\ref{sec:problem}, the pipeline is thus the following: 
\begin{enumerate} 
\item We generate a finite-size CHMM, $(\boldsymbol{F}_s, \boldsymbol{\theta}_s)$ and $(\boldsymbol{F}_t, \boldsymbol{\theta}_t)$.
\item We obtain the weights to be transferred, $\boldsymbol{\hat{w}}_1$, via numerical optimization on the source task. This first computational step is the most demanding in the pipeline.
\item We estimate through Monte Carlo sampling the population covariances $\Omega$ and $\Phi$ that serve as inputs to the asymptotic formulas in equation \eqref{eq:free_entropy}. Note that the computational cost of the MC sampling is minimal.
\item Finally, we iterate the set of saddle-point equations derived in \cite{loureiro2021capturing} from the extremum operation in equation \eqref{eq:free_entropy}, in order to get an analytic prediction for the associated test error. 
\end{enumerate}
Altogether, the steps require small computational power and can be performed in an ordinary laptop in less then 3 minutes, more precisely the second step takes 1-2 minutes, the third 5-10 seconds, and the forth some fractions of a second.

Note that, to the best of our knowledge, it is not possible to circumvent the numerical optimization entailed in step 2 and obtain a fully analytical description of this simple transfer learning setting. This type of analysis would require an analytical characterization of batch-learning settings in two-layer networks, in high-dimensions (both input and hidden-layer dimensions) and in the feature learning regime. This is a highly sought after but still missing piece in the current theory of deep learning.

A further thing to notice is that the replica results are valid in the high-dimensional limit. However, the pipeline proposed in \cite{loureiro2021capturing} and employed in our work, starts from finite-size correlation matrices estimated through Monte Carlo sampling. The justification is the following: the final replica expressions only depend on the traces of the correlation matrices. In high dimensions, the traces translate into expectations over the corresponding spectral distributions. However, these expectations can be well approximated by empirical averages over the finite dimensional spectra, provided the dimension is large enough (we employ $H\sim 1e3$ in our simulations). 

We employ the same analytic framework also for the characterization of the performance of RF, for which the GET assumption was recently proven rigorously \cite{hu2020universality}. In the following, we use the performance of the RF as a baseline for evaluating the effect of transfer learning in the CHMM model. For further comparisons, in the following we also consider: the performance of a two-layer network (2L) trained from scratch on the target task; and the effect of transfer learning combined with an additional fine-tuning stage of the entire network (ft-TF). Both these performances are obtained numerically. The details on the hyper-parameters employed in the learning protocols are provided in \ref{app:simulation_details}.

\section{Results}\label{sec:result}

We now employ the CHMM and the analytic description of the TF performance to explore the effectiveness of transfer learning in a controlled setting. For simplicity, we focus our presentation on two key variables that can impact the generalization of TF. We first consider the effect of different correlation levels between source and target tasks, and afterwards we examine the role played by the latent dimensions of the two datasets. 
In both cases the starting point is an experiment on real data, followed by the identification of similar phenomenology in the synthetic model. Legitimized by the observed qualitative agreement, we then use the described toolbox of analytical and numerical methods to explore associated transfer learning scenarios and draw corresponding phase diagrams. 

\subsection{Transfer between datasets with varying levels of correlation.} A highly intuitive aspect of transfer learning is that the degree of task relatedness crucially affects the results of the transfer protocol. As mentioned in section~\ref{sec:problem}, many sources of correlation can relate different learning tasks, creating a common ground for feature sharing among neural network models.

We start with a simple transfer learning experiment on real data. We consider the EMNIST-letters dataset \cite{cohen2017emnist}, containing centered $28\times 28$ images of hand-written letters. We construct the source dataset, $\mathcal{D}_s$, by selecting two groups of letters ($\{A,B,E,L\}$ in the first group and $\{C,H,J,S\}$ in the second) and assigning them binary labels according to group membership. We then generate the target task, $\mathcal{D}_t$, by substituting one letter in each group (letter $E$ with $F$ for the first group and letter $J$ with $I$ for the second). In this way, a portion of the input traits characterizing the source task is replaced by some new traits in the target.

\begin{figure}[]
    \centering
    \begin{subfigure}[b]{0.49\textwidth}
        \includegraphics[width=\linewidth]{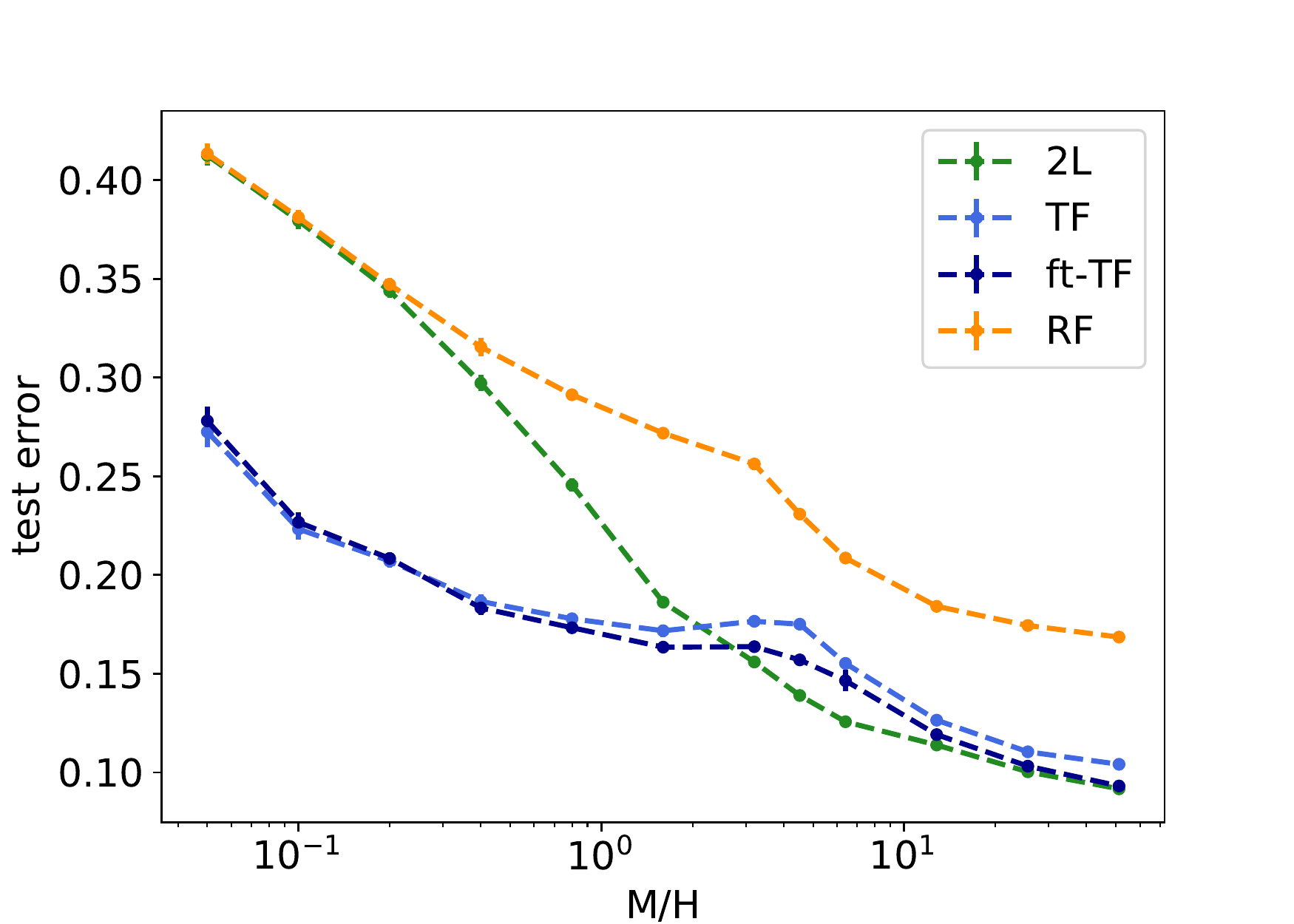}
        \caption{real data}
    \end{subfigure}
    \begin{subfigure}[b]{0.49\textwidth}
        \includegraphics[width=\linewidth]{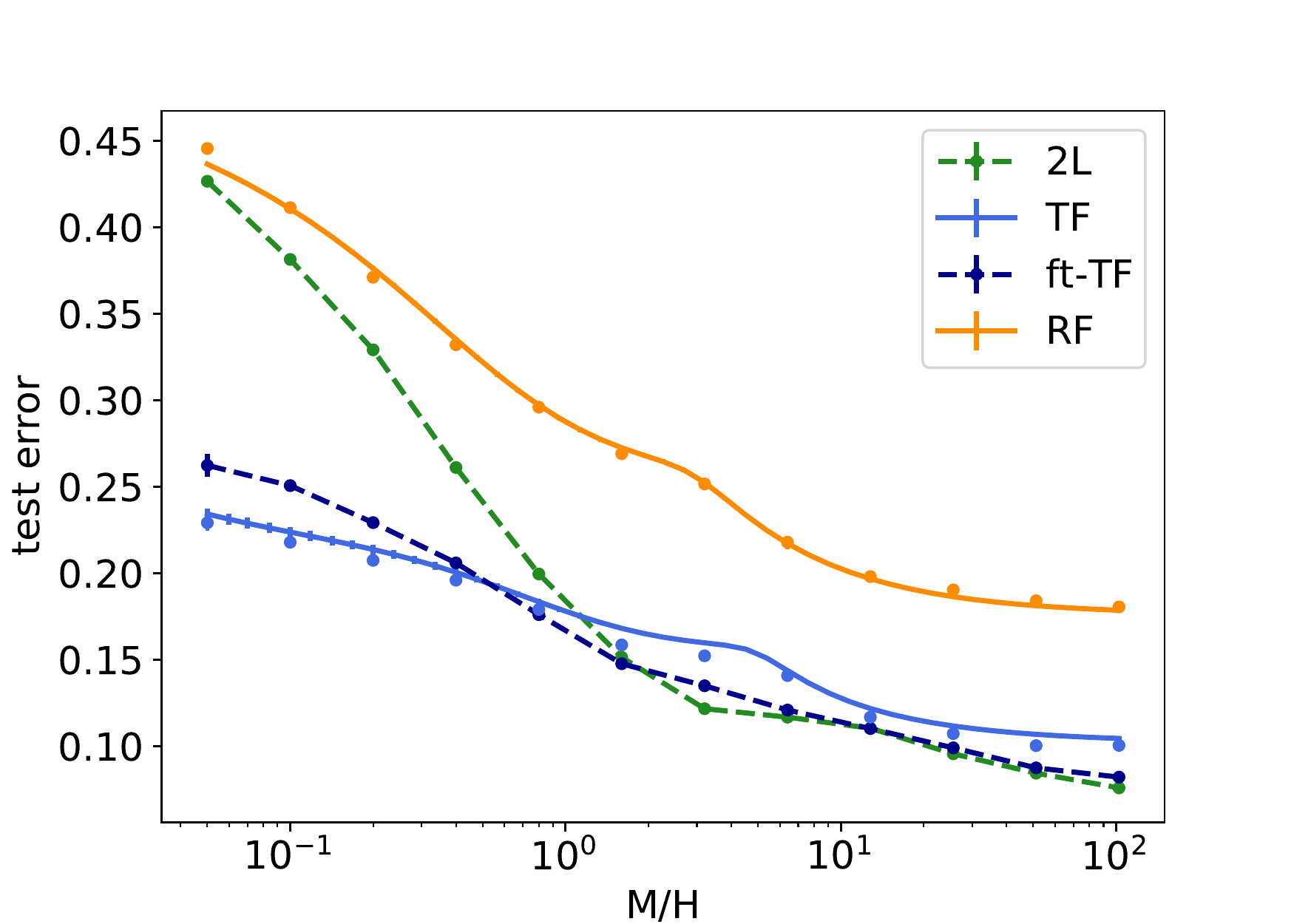}
        \caption{synthetic data}
    \end{subfigure}
    \caption{\textbf{Role of feature perturbation}: test error as a function of the number of samples per hidden unit for both (a) Transfer between two subsets of EMNIST-letters. (b) CHMM with $30\%$ of substituted generative features in the target task. Full lines are from the theoretical predictions for the CHMM, points refer to numerical simulations with connecting dashed lines. The parameters of the networks are set to: $H = 500$, $\lambda = 1e-8$ on real data and $\lambda = 1e-7$ on synthetic data. The input sizes are set to $D = 784$ for the experiments with real data and $D = 1000$ for those with the synthetic model. The parameters of the synthetic data model are set to: $L_{s}=L_{t} = 150$, $\left(\eta, \rho, q \right) = \left( 1, 0.3, 1\right)$. The size of the source task is $M_s = 25600$ on real data and $M_s = 51200$ on synthetic data. All the experiments have been averaged over 50, 20 and 10 samples for low, intermediate and high numbers of samples respectively. In all figures, the error bars are smaller than the size of the points.}
    \label{fig:teacher_pert_real_vs_synthetic}
\end{figure}

Fig.~\ref{fig:teacher_pert_real_vs_synthetic}(a) displays the results. On the x-axis, we vary the number of samples in the target task (notice the log-scale in the plot), while the size of the source dataset is large and fixed. A first finding is that TF (light blue curve) consistently outperforms RF (orange curve), demonstrating the benefit of the feature map transfer. More noticeably, at a low number of samples, TF substantially outperforms training with 2L (green curve), with test performance gains of up to $15\%$. This gap closes at intermediate number of samples, and we observe a transition to a regime where the 2L performs better. 
We also look at the effect of fine-tuning the feature map in the TF, finding that at small numbers of samples it is not beneficial with respect to TF, while at large numbers of samples it modestly helps generalization. Both the learning curves of TF and RF display an interpolation peak, due to the low employed regularization. This type of phenomenon -- connected with the double-descent phenomenon -- has sparked a lot of theoretical interest in recent years \cite{belkin2019reconciling,opper1996statistical}. The cusp is delayed in the TF, signaling a shift in the linear separability threshold due to the transferred feature map. It is also interesting that the fine-tuned TF retains the cusp while the 2L model does not. This is due to the different initialization: while 2L starts from small random weights, ft-TF picks up the training from the TF, where the second-layer weights can be very large. Further details are provided in section \ref{sec:separability_threshold}.

While the reported empirical observations are not too surprising, we can now see if similar behaviors can be traced also in a synthetic setting. We can straightforwardly reproduce the type of dataset correlations described above via feature substitution in the CHMM, as described in section \ref{sec:problem}. Fig.~\ref{fig:teacher_pert_real_vs_synthetic}(b) shows the transfer learning phenomenology in the CHMM, displaying a remarkable similarity with the previous experiment. In the plot, the full lines show the results of the theoretical analysis, corroborated by numerical simulations in finite-size (points in the same color). The observed agreement between theoretical predictions and numerical experiments validates the GET assumption behind the analytic approach, as described in section~\ref{sec:theory}. The 2L and ft-TF dashed lines are instead purely numerical, averaged over different realizations of the CHMM. For low number of samples we observe rather clearly that the fine-tuning of TF actually deteriorates the test error due to over-fitting, in accordance with empirical findings in \cite{yosinski2014transferable}. This effect is also seen in another real data experiment reported in \ref{app:exp_real_data}.

The correspondence between the transfer learning behavior observed on real data and in the CHMM motivates a more systematic exploration of the possible transfer learning regimes using the explicit fine-grained control over the dataset correlations in the CHMM.
The analytical pipeline presented in section \ref{sec:theory} is extremely efficient for deriving phase diagrams in the CHMM. In particular, with a single numerical optimization for $\boldsymbol{\hat{w}}_1$ and a single MC sampling for the population covariance $\Omega$, we can obtain an entire slice of the phase diagram at fixed source dataset, via repeated iterations of the asymptotic equations. As a result, an entire phase diagram (with about 1000 points, 10 samples per point) can be derived in 5-10 hours on a modern laptop (compared with order $\sim10^3$ hours for obtaining the same diagram through pure numerical optimization). 

\begin{figure}[]
    \centering
    \begin{subfigure}[b]{0.325\textwidth}
        \includegraphics[width=\linewidth]{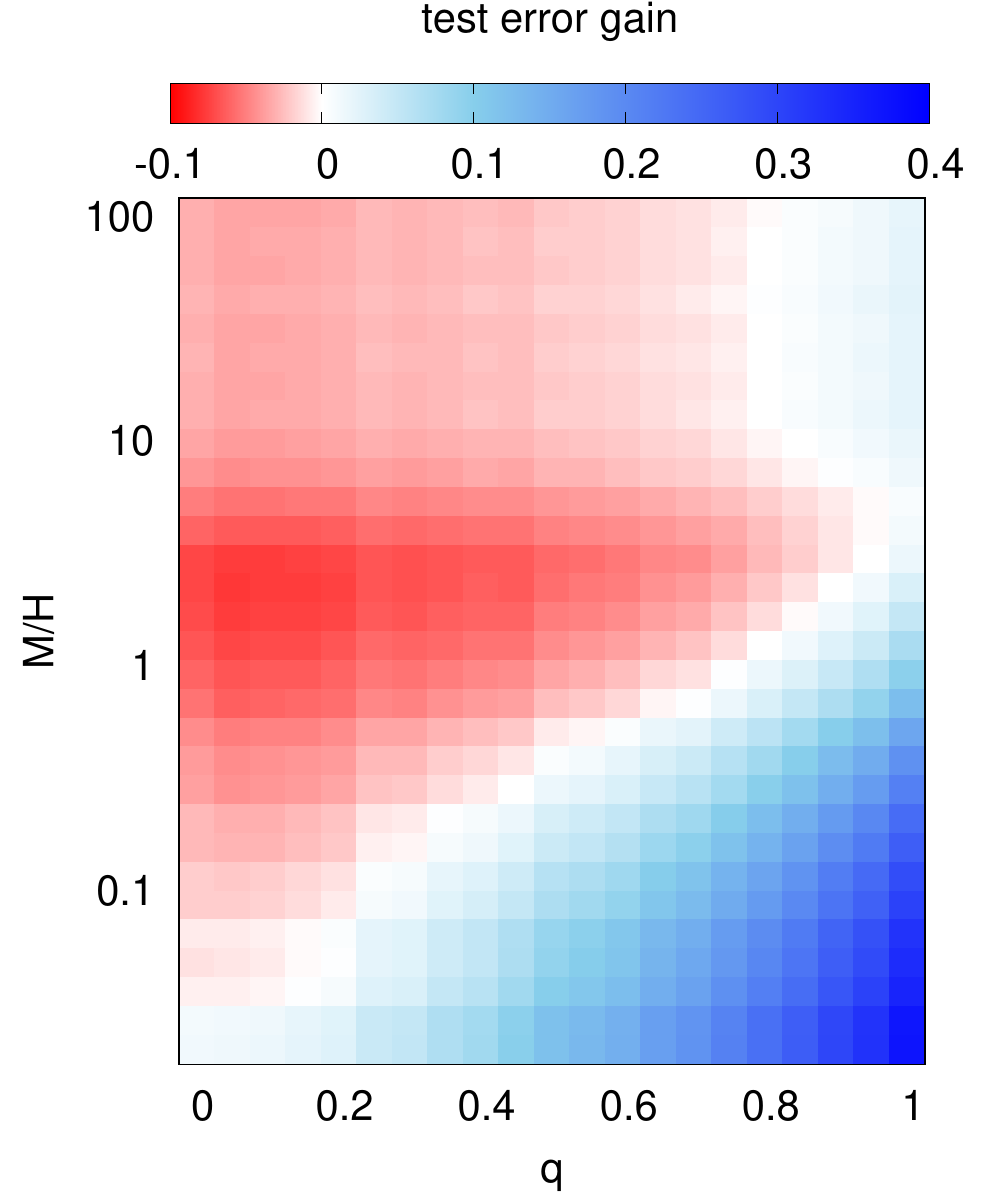}
        \caption{TF to 2L gain}
    \end{subfigure}
    \begin{subfigure}[b]{0.325\textwidth}
        \includegraphics[width=\linewidth]{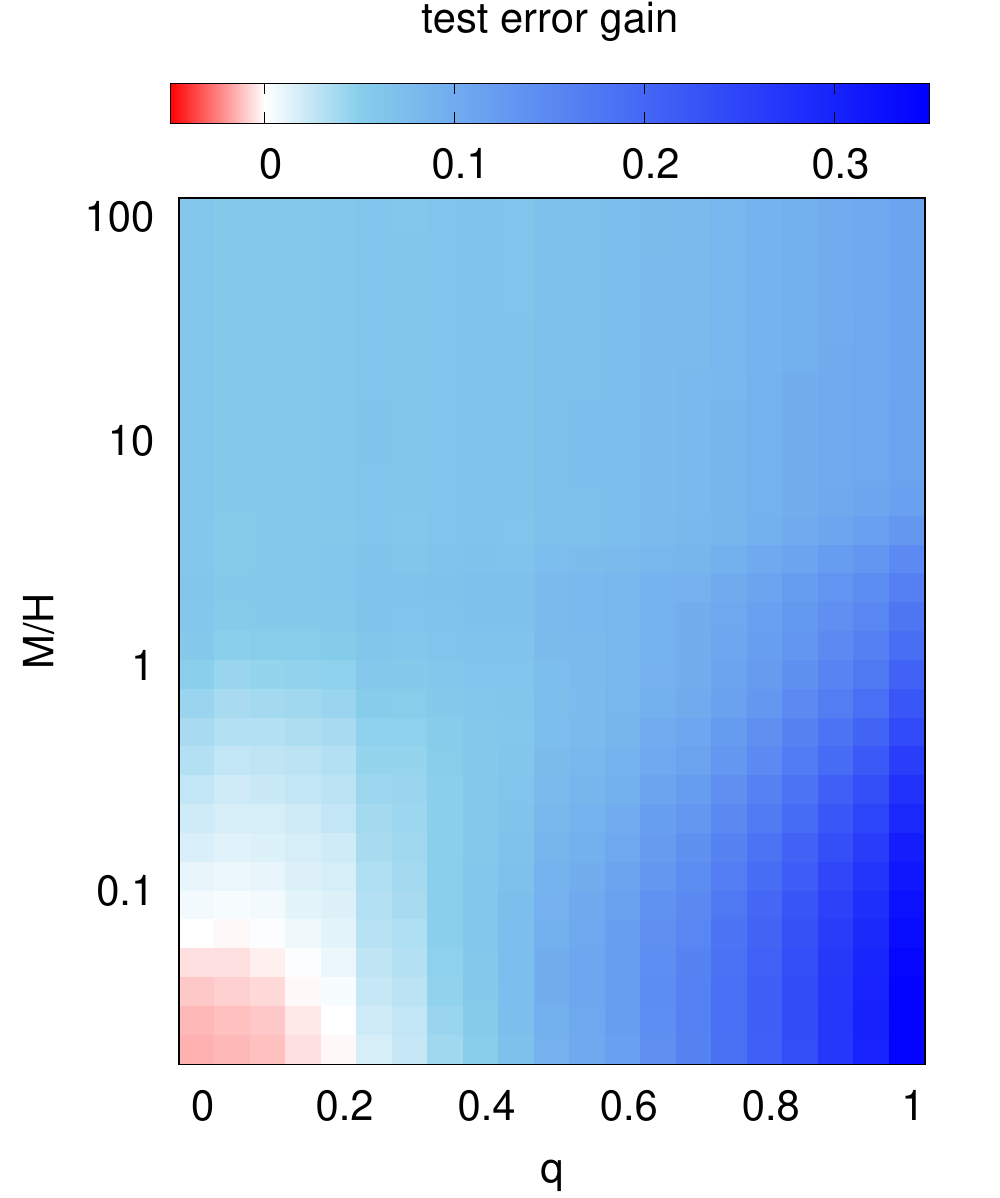}
        \caption{TF to RF gain}
    \end{subfigure}
    \begin{subfigure}[b]{0.325\textwidth}
        \includegraphics[width=\linewidth]{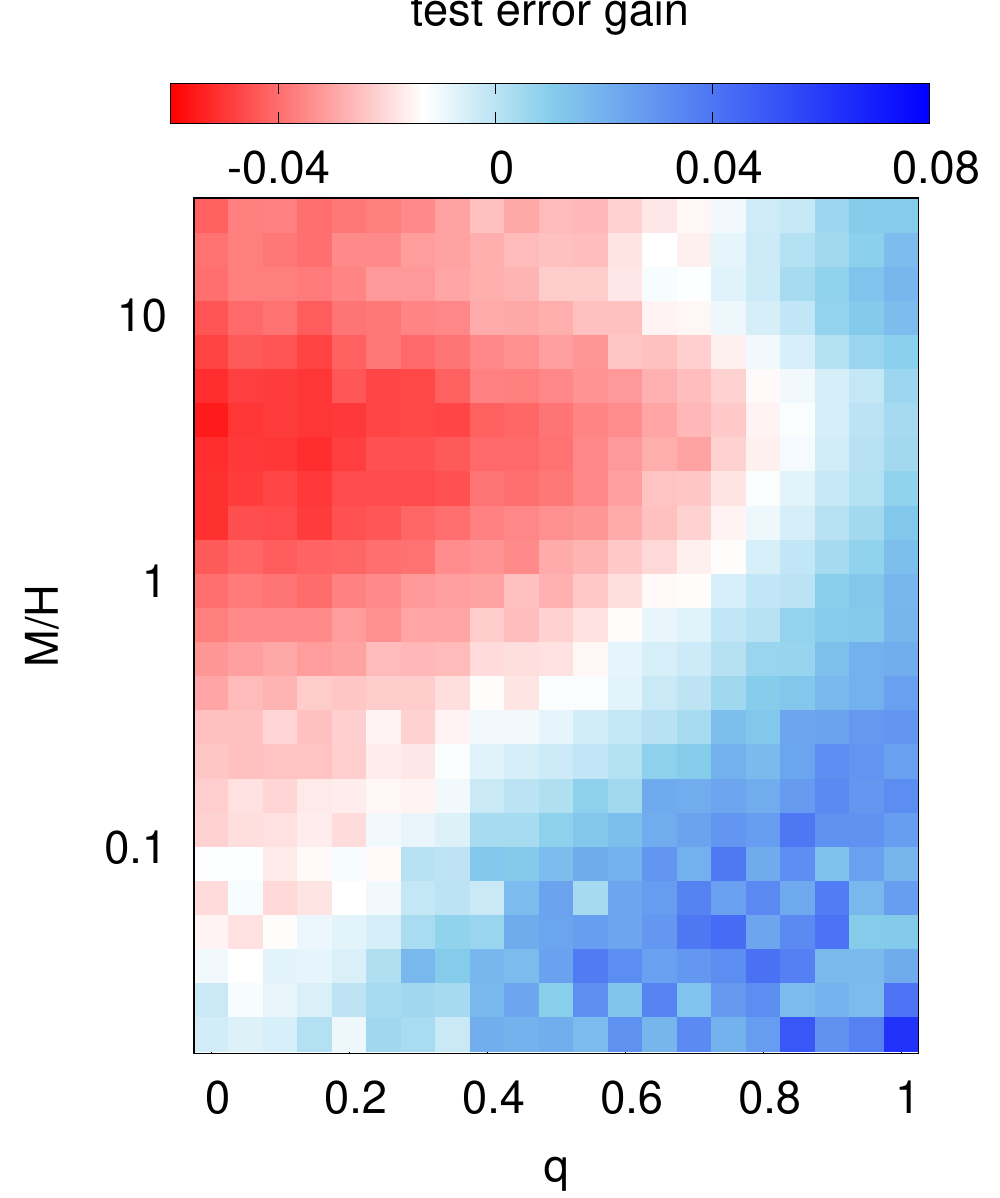}  
        \caption{TF to ft-TF gain}
    \end{subfigure}
    \caption{\textbf{Effect of the dataset correlation on transfer learning performance}. The three phase diagrams characterize the learning performance in the CHMM, comparing the transferred feature model (TF) with: (a) a two-layer network with both layers trained on the target task (2L), (b) a random feature model trained on the same task (RF), and (c) the transferred feature model after an additional stage of fine-tuning (ft-TF), as a function of the correlation parameter~$q$ and the target dataset size $M$. Blue indicates TF performing better than the compared algorithm (conversely, red indicates TF performing worse). The results for TF and RF are obtained from the theoretical solution of the CHMM, while 2L and ft-TF are purely numerical ($10$ samples per point). The parameters of the model were set to $L=200$, $(\eta, \rho)=(1,0)$, $D=1000$ and $H=500$.}
    \label{fig:teacher_pertubations_phase_diagram}
\end{figure}

As a representative case, Fig~\ref{fig:teacher_pertubations_phase_diagram} shows three phase diagrams, comparing (a) TF with 2L, (b) TF with RF, (c) and ft-TF with TF, and displaying the performance gain as the teacher network alignment $q$ is varied. High values of $q$ indicate strongly correlated tasks, while low values indicate unrelated ones. Each point in the diagrams represents a different source-target pair, with a variable number of samples in the target task. In panel (a), we can see that TF can outperform 2L (blue shaded region) when the number of samples is low enough, or when the level of correlation is sufficiently strong. In these regimes 2L is not able to extract features of a comparable quality. Note that, at $q=1$, the transferred features are received from a source task that is identical to the target, so it is to be expected that at high numbers of samples the performance of TF is equivalent to 2L. The darker red region (around numbers of samples per hidden unit of order 1) is connected to the appearance of the interpolation peak, mentioned above and investigated further in section \ref{sec:separability_threshold}.
In panel (b), we see a corner of the phase diagram (red shaded region) where the performance of TF is sub-obtimal with respect to learning with random features. This is a case of negative-transfer: the received features are over-specialized to an unrelated task and hinder the learning capability of TF as in \cite{rosenstein2005transfer}. Finally, in panel (c), we can see when adding a final stage of fine-tuning on the new data (adapting the weights in both layers) can induce better generalization performance. The red shaded region, highlighting this improvement, is found at lower correlations (where the transferred features are not effective without fine-tuning) and higher number of training samples in the target dataset (where additional training does not lead to over-fitting). At small number of samples, instead, this procedure leads to over-fitting \cite{yosinski2014transferable}. 

We here reported on the effect of a teacher network perturbation, but we find surprisingly similar results for the other families of transformations, perturbing the generative features with $\eta$ and $\rho$ (the corresponding phase diagrams are shown in \ref{app:exploring_parameters}). Thus, at the level of the CHMM, all these manipulations have a similar impact on dataset correlation and feature map transferability. Next, we asked whether these patterns held qualitatively on real world data. Remarkably, we trace similar trends in different transfer learning settings on MNIST-like datasets, as documented in \ref{app:exp_real_data}. 
We emphasize that obtaining almost identical behavior in experiments on MNIST-like datasets and in a simple (fundamentally Gaussian) synthetic model is non-trivial, and suggests that our setting captures important features of the real data setting. The fact that transfer learning performance is largely insensitive to the minute details of source and target datasets, and seems instead to be dominated by the intensity of the dataset correlations, hints at the existence of a class of universality connecting apparently disparate learning problems. In this perspective, although the obtained phase diagrams cannot be practically employed for predicting real data behavior, the uncovered phases and lines between them are likely to delineate a picture which is robust to moderate changes in the data-model. In sum, our systematic phase space analysis suggests the following conclusions: transfer learning helps most in the limited target data, high task correlation regime; sufficiently dissimilar tasks can cause negative transfer; and fine tuning is only beneficial given sufficient target task data.

\paragraph{Connection with related works.} Several cross-domain transfer learning experiments have been carried out in the past decades. For example, \cite{kornblith2019better} transferred from the imagenet dataset to different target sets, but without investigating the relatedness between target and source, while  \cite{peng2019moment} extensively studied transfer learning from/to several sources/targets. These experiments generally show that as the difference between source and target datasets increases the benefit from transferring decreases, as observed in our study. Similar results are described in \cite{neyshabur2020being}, where the transfer is considered from imagenet \cite{krizhevsky2012imagenet} to different kind of images (medical \cite{irvin2019chexpert}, pictures, drawings \cite{peng2019moment}). 
The \textit{negative transfer} effect \cite{rosenstein2005transfer} -- i.e. a disadvantage in employing transfer learning with respect to training from scratch -- found in a corner of Fig.~\ref{fig:teacher_pertubations_phase_diagram} has also been observed in recent numerical works on transfer learning. This phenomenon can be caused by problems in the architecture \cite{yosinski2014transferable}, or can be associated to task-related difficulties. For example, \cite{merkow2017deepradiologynet} observed negative transfer from imagenet to tomography head scans, imputing the poor performance to the datasets dissimilarity, while \cite{zhang2020overcoming} reviewed negative transfer in applications where the datasets belong to far-apart domains.

On a different note, there have been few proposals of viable metrics for evaluating similarity between real datasets, with the goal of obtaining predictors of the associated transfer learning performance. Some of these architecture agnostic metrics evaluate dataset distances based on information theory, information geometry and optimal transport \cite{tran2019transferability,alvarez2020geometric,nguyen2021similarity,gao2021information,achille2021}. An interesting direction for future work would be to connect these distance metrics with the parametric transformations in the CHMM.

\subsection{Transfer between datasets of different complexity.}\label{sec:asymmetric}
So far, we discussed situations where the transfer occurs between tasks with similar degrees of complexity. However, in deep learning applications, this is not the typical scenario. Depending on whether the source task is simpler or harder than the target one, a different gain can be observed in the generalization performance of TF. The two transfer directions, from hard to simple and vice-versa, are not symmetric. This observation has been repeatedly reported for transfer learning applications on real data \cite{liu2015simple,menegola2016towards,neyshabur2020being}.  

\begin{figure}[]
    \centering
    \begin{subfigure}[m]{0.14\textwidth}
        \vspace{-0.65cm} hard$\to$easy \\ \\ \\ \\ \\
        easy$\to$hard
    \end{subfigure}
    \begin{subfigure}[m]{0.42\textwidth}
        \includegraphics[width=\linewidth]{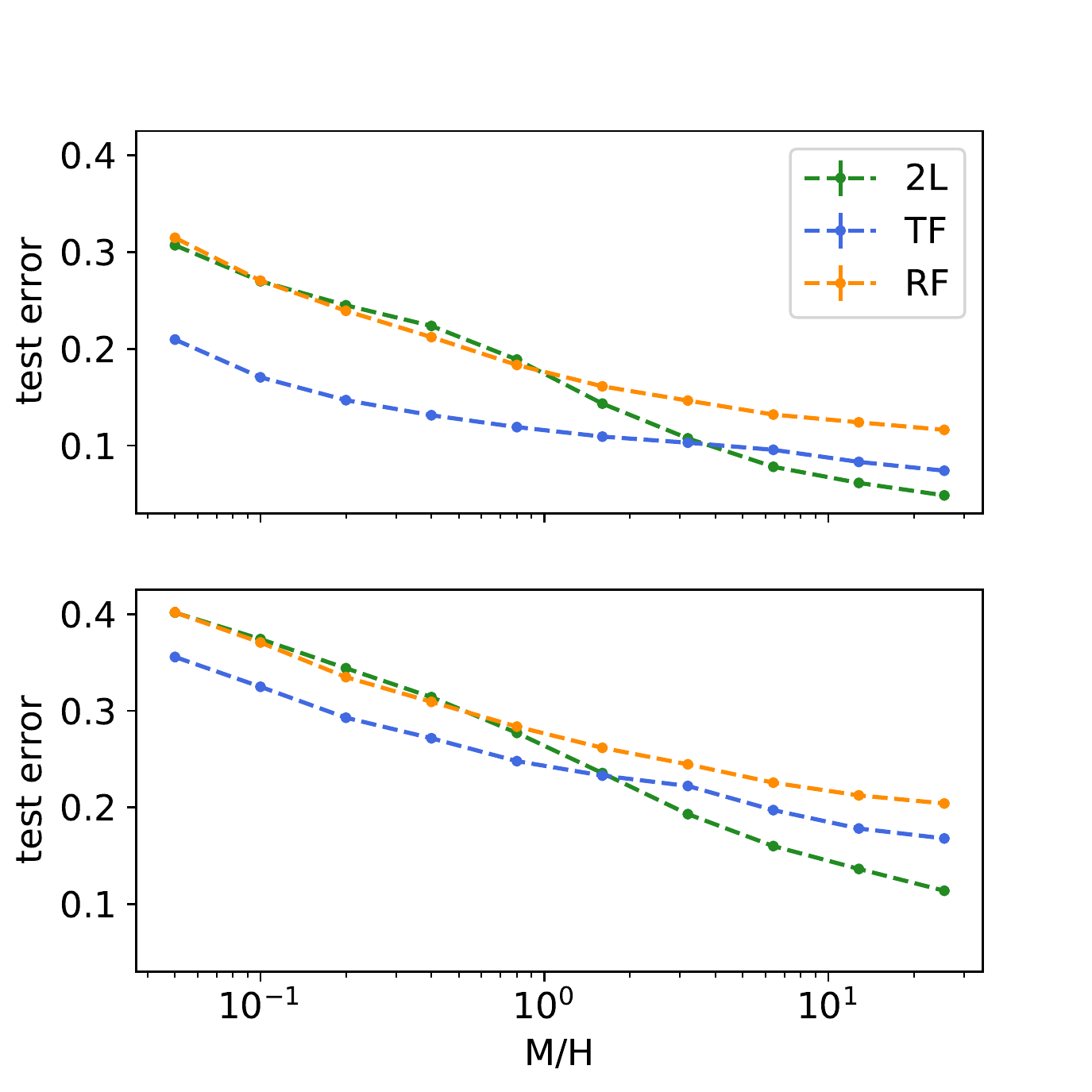}
        \caption{real data}
    \end{subfigure}
    \begin{subfigure}[m]{0.42\textwidth}
        \includegraphics[width=\linewidth]{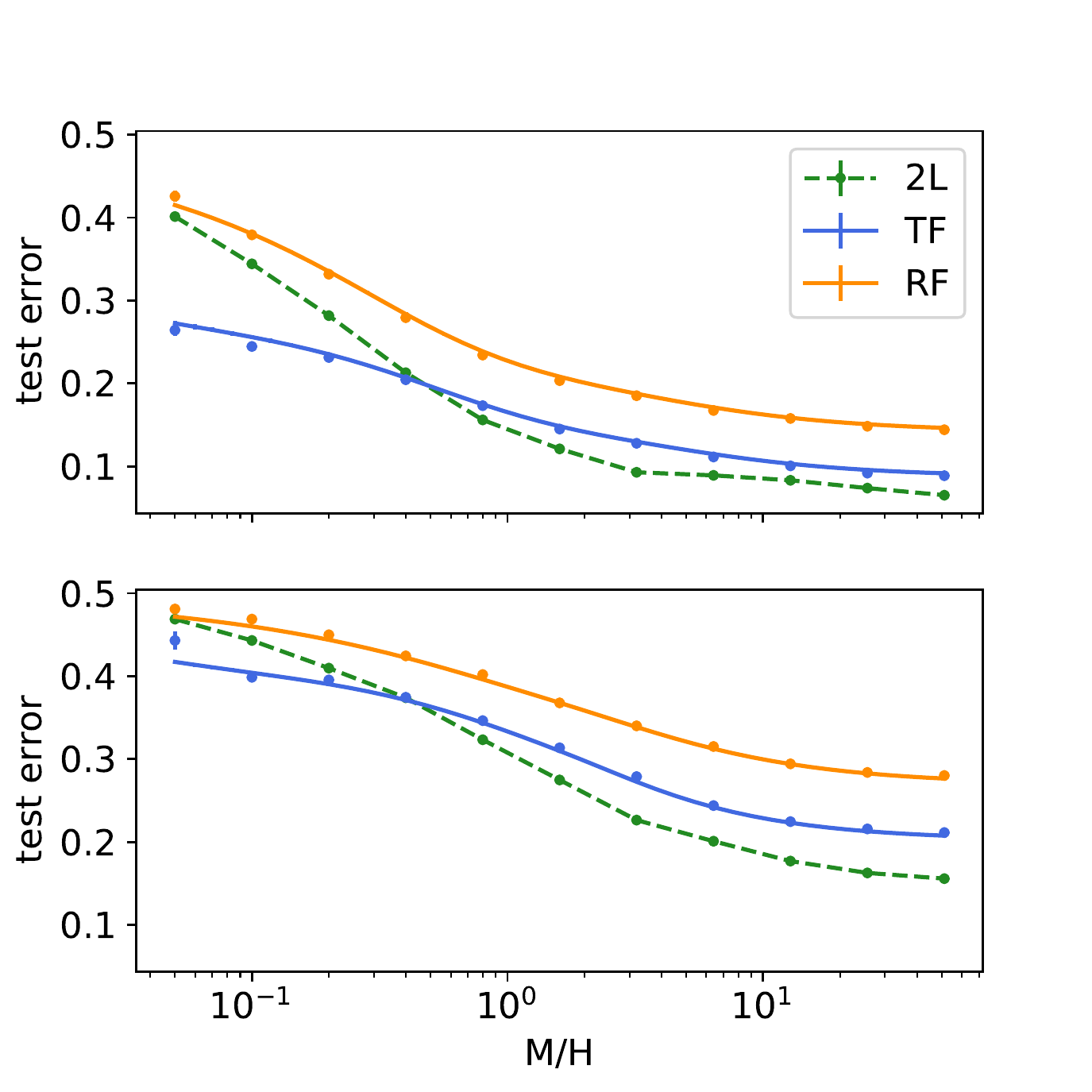}
        \caption{synthetic data}
    \end{subfigure}
    \caption{\textbf{Asymmetric transfer}: test error as a function of the number of samples per hidden unit for both (a) real and (b) synthetic data. Top plots show the hard-to-easy transfer direction, bottom plots the easy-to-hard transfer direction. (a) Full EMNIST-letters vs. filtered EMNIST-letters. (b) CHMM with different latent dimensions of the source and target model. Full lines refer to theoretical predictions for the CHMM, points refer to numerical simulations with dashed lines being a guide to the eye.  
    The parameters of the networks are set to: $\lambda = 1e-6$, $H = 500$. The input size is $D = 784$ for the experiments on real data and $D = 1000$ for the synthetic model. The parameters of the synthetic data model are set to: $L_{\text{easy}}=100$ and $L_\text{hard}=400$, $\left(\eta, \rho, q \right) = \left( 0.8, 0, 0.9\right)$. The size of the source task is $M_s = 24000$ for real data and $M_s = 51200$ for synthetic data in both transferring directions (hard-easy and easy-hard). All the experiments have been averaged over 50, 20 and 10 samples for low, intermediate and high numbers of samples respectively. In all figures, the error bars are smaller than the size of the points.}
    \label{fig:asymmetric_real_vs_synthetic}
\end{figure}

To isolate the asymmetric transfer effect, we propose again a simple design. Consider as a first classification task, $\mathcal{D}_\text{hard}$, the full EMNIST-letters dataset (including all classes) with binarized labels (even/odd categories). As a second task, $\mathcal{D}_\text{easy}$, consider instead a filtered EMNIST dataset (containing only some of the letters) with the same binarized labels. As denoted by the subscript, the learning problem associated to the first task is harder, given the richness of the dataset, while the second classification task is expected to be easier.  Fig.~\ref{fig:asymmetric_real_vs_synthetic}(a) shows the outcome of this experiment. In the top sub-plot, we display the transfer from $\mathcal{D}_{\text{hard}}$ to $\mathcal{D}_{\text{easy}}$, while in the lower sub-plot the transfer from $\mathcal{D}_{\text{easy}}$ to $\mathcal{D}_{\text{hard}}$. A first remark is on the different difficulty of the two learning problems: as expected, the test error is  smaller in the top figure than in the bottom one, especially when the number of samples is small (difference of about $10 \%$ asymptotic test accuracy for all learning models). However, a more surprising observation is that, with few target samples, the performance gain of TF over the two base-lines is not symmetric in the two transfer directions: the gain is large when the transfer goes from $\mathcal{D}_{\text{hard}}$ to $\mathcal{D}_{\text{easy}}$ (top figure, about $10 \%$), while it is smaller in the opposite transfer direction (bottom figure, about $5\%$).

As mentioned in section \ref{sec:problem}, dataset complexity is captured in our modeling framework through the latent dimension of the two HMMs. In particular, we can respectively assign a higher $L_\text{hard}$ and a smaller $L_\text{easy}$ to the two tasks. Correspondingly, the harder HMM will comprise a larger number of generative features. Fig.~\ref{fig:asymmetric_real_vs_synthetic}(b) shows the asymmetric transfer effect in the setting of the CHMM. 
Again, the top sub-plot shows the transfer from $L_\text{hard}$ to $L_\text{easy}$, while the bottom sub-plot the converse. The different task complexity is reflected in the lower test scores recorded when the target task is the easier one. Moreover, as above, we observe a different gain between the two transfer directions for TF (above $10 \%$ in transfer from hard to easy, about $5\%$ in the other direction). Thus, the difference in the intrinsic dimension of the two datasets seems to be the key ingredient for tracing this phenomenon.

We can now exploit our modeling framework and further explore the asymmetric transfer effect as a function of the latent dimension discrepancy. We consider source and target HMMs with variable latent dimensions, $L_s$ and $L_t$, while keeping $L_s + L_t = 500$ fixed. This allows us to probe cases where the target task is simpler and cases where it is more complex than the source. By comparing the performance of TF to RF, we identify the regimes where transfer learning produces the largest gains. 

\begin{figure}[H]
    \centering
    \begin{subfigure}[b]{0.325\textwidth}
        \includegraphics[width=\linewidth]{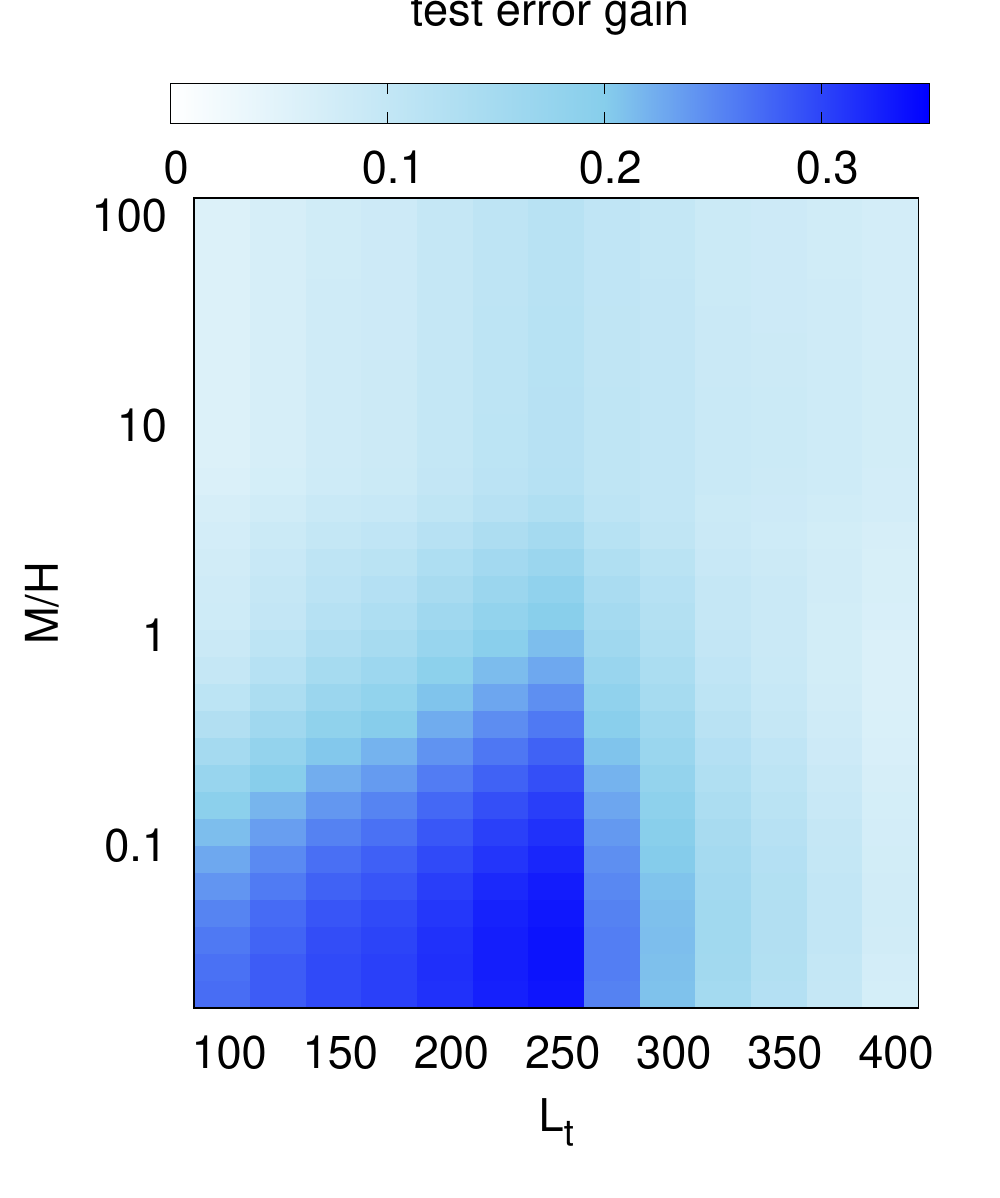}
        \caption{TF to RF gain}
    \end{subfigure}
    \begin{subfigure}[b]{0.55\textwidth}
        \includegraphics[width=\linewidth]{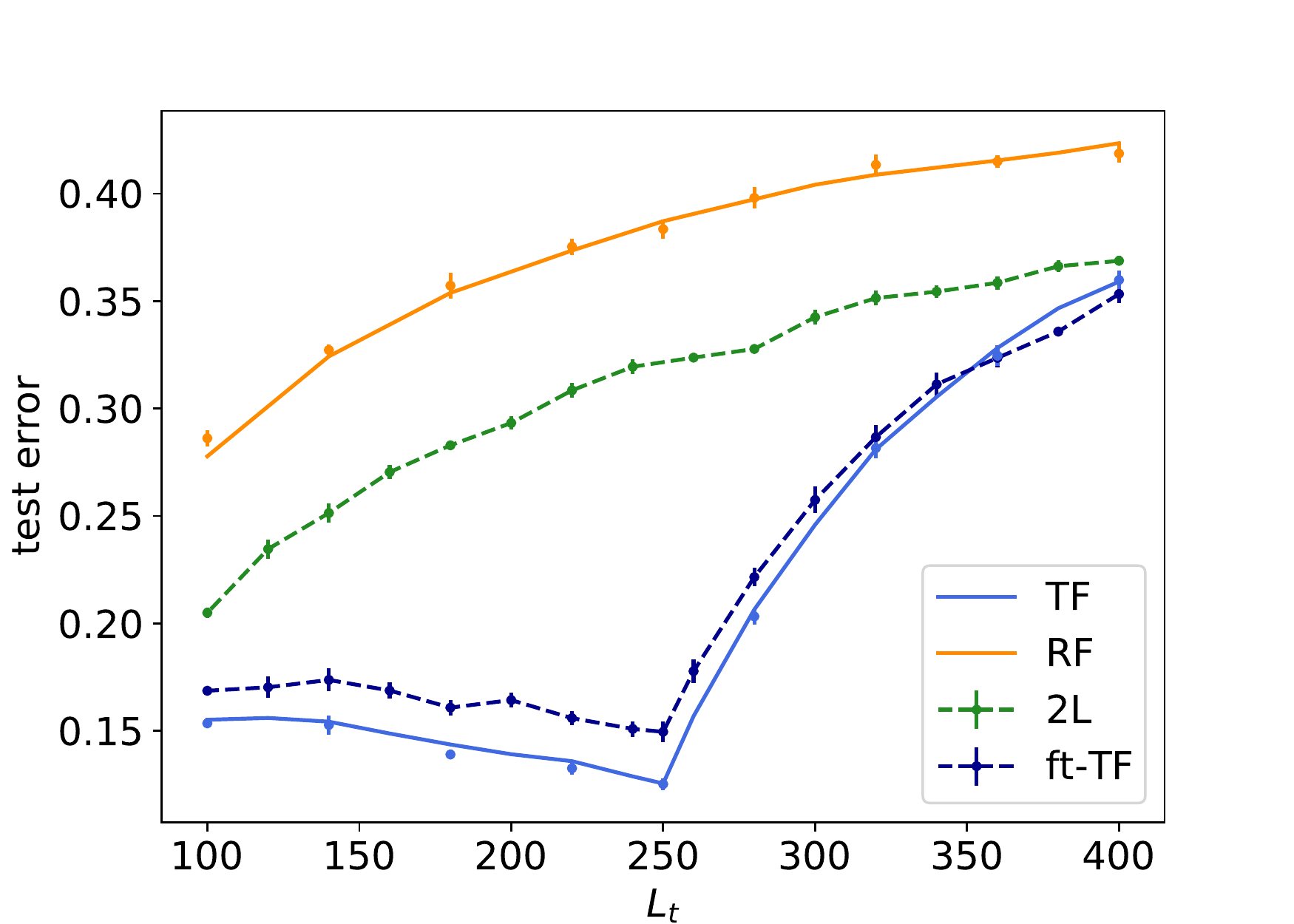}
        \caption{horizontal slice of the phase diagram}
    \end{subfigure}
    \caption{\textbf{Effect of a latent dimension asymmetry}. (a) Phase diagram comparing the performance of the transferred feature model (TF) to that of a random feature model (RF), as a function of the latent dimension of the target task $L_t$ (larger $L$ implies higher dataset complexity) and of the number of samples in the target dataset. In the source task we fix $L_s=500-L_t$, therefore the vertical axis at $L_t=250$ indicates the case of symmetric latent spaces. The displayed results are obtained from the theoretical solution of the CHMM. (b) Horizontal slice of the phase diagram, at $M/H=0.4$, in which also the performance of the two-layer network (2L) and the fine-tuned TF are shown. Full lines are obtained from the theoretical solution of the CHMM, points from numerical simulations, dashed lines are guide to the eye (10 seeds per point). The parameters of the generative model are set to $(\eta,\rho,q)=(1,0,1)$, $D=1000$ and $H=500$, $\lambda=1e-6$.}
    \label{fig:asymmetric_phase_diagram}
\end{figure}

Fig.~\ref{fig:asymmetric_phase_diagram}(a) shows the resulting phase diagram, highlighting a stark asymmetry when transferring between datasets with different latent dimensions. In the plot, the x-axis shows the latent dimension of the target task, therefore the hard to easy regime is found on the left, while the easy to hard regime is found on the right. The vertical axis at $L_t=250$, represents the symmetry line of the phase diagram and corresponds to the case where $L_s=L_t$, namely when the two HMMs share the same generative features. By moving to the right or to the left of the axis by the same amount, the number of generative features common to both datasets is identical. However, on the right, the target dataset is more complex than the source dataset. As a result, on this side of the phase diagram, the performance gain of TF is smaller. When instead the target task is simpler, the test performance of TF is highly improved, especially at low numbers of samples.

Fig.~\ref{fig:asymmetric_phase_diagram}(b) displays a horizontal cut of the phase diagram, at $M/H=0.4$. In the low number of samples regime transfer learning is highly beneficial, and indeed we observe that TF greatly outperforms RF and 2L. Again, note that at $L_t=250$ the latent dimensions of source and target tasks are equal. However, as we move away from the symmetry line at $L_t=250$, we see the asymmetric transfer effect coming into play. When the target latent dimension increases, the gap in performance closes faster. This is due to the lack of learned features, affecting the test error of TF. If we look at the effect of fine-tuning the transferred features, we see a different behavior at small/large target latent dimension. When $L_t$ is small, ft-TF is detrimental. When $L_t$ is instead large, ft-TF can slightly improve performance. The fully trained 2L network is over-fitting due to the small dataset.

\paragraph{Connection with related works.} 
Although in practice it is common to transfer from non-target-specific datasets to the target dataset (e.g. \cite{yosinski2014transferable,howard2018universal}), to the best of our knowledge, a systematic exploration of the generalization performance from complex (feature rich) to simple (few features) has not being carried out. However, we can obtain an indication of this effect in the already cited works \cite{peng2019moment,neyshabur2020being} on generalization performance with different source and target sets, by carefully comparing the reported scores (for instance in  \cite{peng2019moment} table 3 we see that transferring from real images to paintings achieve better performance than black and white sketches to paintings). Our results help systematize numerous observations of the value of adapting large models trained in rich settings, a strategy which has proven useful in a range of settings.

\subsection{Separability threshold and double descent.}\label{sec:separability_threshold} 

Conventional wisdom in machine learning and statistical learning theory says that the generalization performance of learning models is affected by the so-called bias-variance tradeoff. When the number of parameters in the learning model is too small, the resulting estimator may underfit the training set, thus failing in correctly predicting the correct input-target relationship (high bias-low variance regime). On the contrary, when the functional space of possible predictors is too large, the selected estimator may overfit the training set, ending up to interpolate the noise in the data samples (low bias-high variance regime). The generalization error then follows the classical U-shaped curve, whose minimum coincides with the ``sweet spot'' balancing these two regimes. 

This picture is at odds with modern deep-learning applications, such as neural network models, for which optimal generalization performances are often attained in the over-parametrized regime. In this case, the generalization error follows the classical U-shaped curve only up to the interpolation threshold, namely when the model can perfectly fit the training data. Beyond the interpolation threshold, the generalization error decreases monotonically with the amount of parameters exploited in the learning model. Between these two regimes, a peak occurs for vanishing regularization. 

This double-descent behavior of the generalization error was first observed in \cite{lecun1991second, krogh1992simple, opper1996statistical} and has recently risen a lot of theoretical interest. In particular, in \cite{belkin2019reconciling} the authors established the existence of the double-descent behavior for a wide-variety of learning models, including neural networks, random feature models and decision trees. The double descent curve has been exemplified in \cite{spigler2019jamming} by means of the well-know jamming phenomena in statistical physics literature, and it has been shown to be crucially affected by the additive noise in the labels and the weights initialization in \cite{d2020double, ADVANI2020428}. The asymptotic expression of the generalization error in the high-dimensional limit (see sec. \ref{sec:theory}) of random feature models has been derived in \cite{mei2019generalization} for ridge regression and in \cite{gerace2020generalisation} for classification. In particular, \cite{gerace2020generalisation} has shown that the double-descent peak occurs at the linear separability threshold of the training data with logistic loss, extending the results in \cite{candes2020phase}.    

\begin{figure}[H]
    \centering
    \includegraphics[width=1.\linewidth]{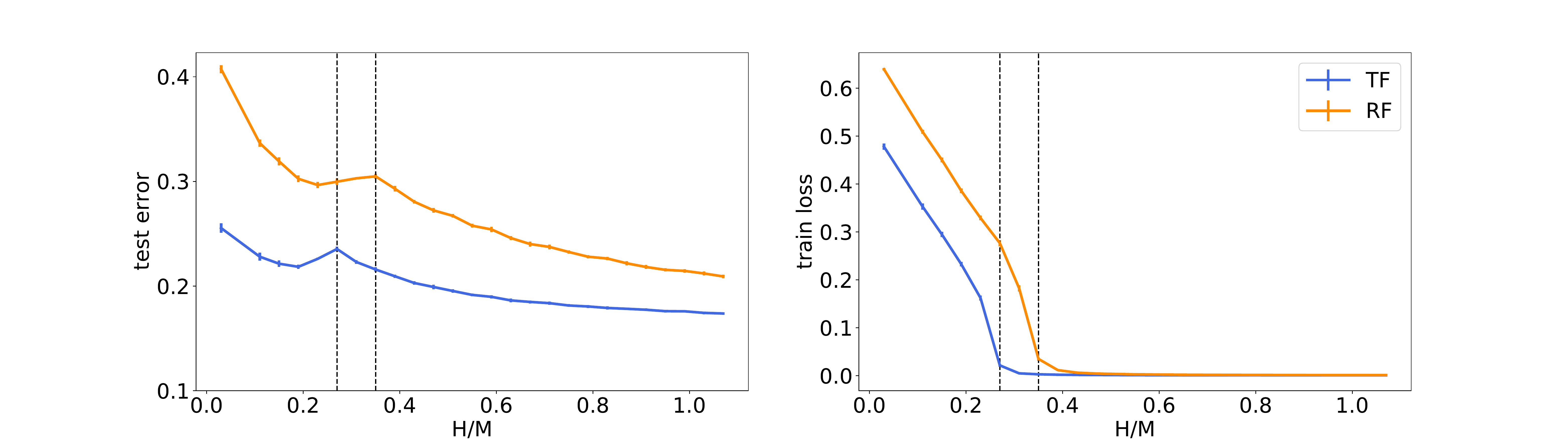}
    \caption{Generalization error (left) and training loss (right) as a function of the number of samples per hidden unit for both RF (orange curve) and TF (blue curve). The parameters of the model are $D = 1000$, $M = 1500$, $L = 200$. The regularization strength is fixed to $\lambda = 1e-8$. In particular, for TF the parameters controlling the correlation between the source and the target task are fixed to $(\eta, \rho, q) = (1, 0.2, 0.8)$. The simulations are averaged over $10$ different realization of the CHMM.}
    \label{fig:double_descent_peak}
\end{figure}

Fig.~\ref{fig:double_descent_peak} displays the double-descent phenomenology in the CHMM for the transferred feature model analyzed in the present work. As in the case of random features \cite{gerace2020generalisation}, a peak in the generalization error can be observed in correspondence of the linear separability threshold. This threshold (dashed black vertical lines in Fig.~\ref{fig:double_descent_peak}) signals the transition to the regime where the input data points can be perfectly separated, i.e. where the training loss is exactly equal to zero. As can be observed in the plot, the interpolation threshold associated to TF is shifted towards smaller model complexities. This is a direct consequence of the non-trivial correlations learned from the source dataset and encoded in the feature map. The preprocessing induced by the transferred features helps in the classification process, making data more easily separable with respect to random Gaussian projections.
Note that the sharpness of the transition is controlled by the regularization strength: the smaller is the regularization, the sharper is the transition between the two regimes. Due to numerical instabilities in the convergence of the saddle-point equations, we could not approach regularization strengths smaller than $\lambda=1e-8$. We also note that, if the regularization strength was optimized separately for each dataset size or if early stopping was employed, this would likely reduce or remove the peak. 

\paragraph{Connection with related works.} 
In the context of transfer learning, \cite{dar2020double,dar2021transfer} have shown the presence of double descent peaks in simple models. In particular \cite{dar2021transfer} explores the difference between the effect induced by transfer learning and standard regularization schemes.

\subsection{Additional results.} 
The modeling framework and the efficient pipeline proposed in this work can be used to investigate other facets of transfer learning. In \ref{app:exploring_parameters} we provide some additional results, exploring the impact of the network width~$H$, and of the number of samples in the source task. In the first case, we find that the biggest performance gaps between TF, RF and 2L are observed in the regime where $H/L$ is smaller than one, while in the opposite regime the gain is observed only at small sample-complexity. In the second case, we find that, at fixed dataset correlation, the number of samples in the source task needs to be sufficiently large in order for the transfer to be effective, as expected. 

Finally, in \ref{app:multiclass} we also challenge the one of the limitations of our modeling setup --the restriction to the binary classification case-- by exploring through numerical simulations the impact of considering multiple classes on the transfer learning performance. To this end, we extend our model and introduce the multi-label CHMM, where relational information among different classes can come into play, and repeat the simulations presented in Fig.~\ref{fig:teacher_pertubations_phase_diagram} and Fig.~\ref{fig:asymmetric_phase_diagram}. Our numerical results show that the behavior is qualitatively equivalent to the binary setting, confirming the robustness of the presented analysis.

\section{Conclusions}

We introduced the correlated hidden manifold model (CHMM), a synthetic model of correlated datasets that allows a semi-analytical characterization of the transfer learning performance for 2 layer neural networks. Despite the simplicity of the model, we were able to show that transfer learning in the CHMM is completely aligned with the behavior recorded in experiments on standard MNIST-derivative datasets. The uncovered universality motivated a more extensive exploration of the possible transfer learning regimes in the CHMM by changing the parametric control over the level of correlation between source and target datasets. We introduced an efficient pipeline for obtaining the associated phase diagrams, which considerably amortizes the high cost of numerical optimization of neural networks. The resulting phase diagrams delineate the transition lines between the transfer learning regimes traced in the literature, and can be used to inspect the impact of the many hyper-parameters at play. We find that transfer is most beneficial in the limited target data, high task correlation regime, and that fine-tuning is only worthwhile for larger target datasets. Further, we document a striking asymmetry in transfer, such that transferring from richer tasks to simpler tasks is more beneficial than vice versa. These results help motivate methods which make use of highly complex models trained on diverse source tasks \cite{howard2018universal, bommasani2021opportunities}.    

Despite the remarkable adherence of the CHMM phenomenology to simple transfer learning experiments on real data, there are several limitations to the presented approach. The main one is technical: in high-dimensions we are not aware of any theoretical framework that captures feature learning in full two-layer networks in a form that could then be used to study transfer learning. Therefore, even our analytic curves are obtained on top of results from numerical learning of the features from the source task. The description of learning processes in architectural variants, commonly used in deep learning practice, is even further from the current reach of existing theoretical approaches. An interesting direction for future work would be to find a method for quantifying the amount of correlation between real source and target datasets and to locate the pair in the phase diagrams we obtained.

\section*{Acknowledgments}
    S.S.M. and A.S. acknowledge Oxford University for the hospitality in early phases of this work. 
    We acknowledge funding from the ERC under the European Union’s Horizon 2020 Research and Innovation Program Grant Agreement 714608-SMiLe, and a Sir Henry Dale Fellowship from the Wellcome Trust and Royal Society (grant number 216386/Z/19/Z). A.S. is a CIFAR Azrieli Global Scholar in the Learning in Machines \& Brains programme. 


\section*{References}
\bibliography{bib}
\bibliographystyle{unsrt}

\newpage

\appendix
\numberwithin{equation}{section}
\numberwithin{figure}{section}

\section{Exploring different aspects of transfer learning in the CHMM}\label{app:exploring_parameters}

In the following paragraphs, we present some additional phase diagrams on the phenomenology of transfer learning in the correlated hidden manifold model (CHMM). These results complement the materials presented in the main text and give additional insights on the role played by the various parameters in transfer learning problems.

\paragraph*{Impact of transformations on the generative features.}
We first consider a similar experiment to the one presented in the first paragraph of section~\ref{sec:result}. In section~\ref{sec:result} we showed the phase diagrams for transfer learning when source and target HMMs are linked by a teacher perturbation. Here we explore the effect of the other types of transformations that preserve the dimensionality of the latent space, namely the feature perturbation and the feature substitution transformations.

\begin{figure}[b]
    \centering
    \begin{subfigure}[b]{0.33\textwidth}
        \includegraphics[width=\linewidth]{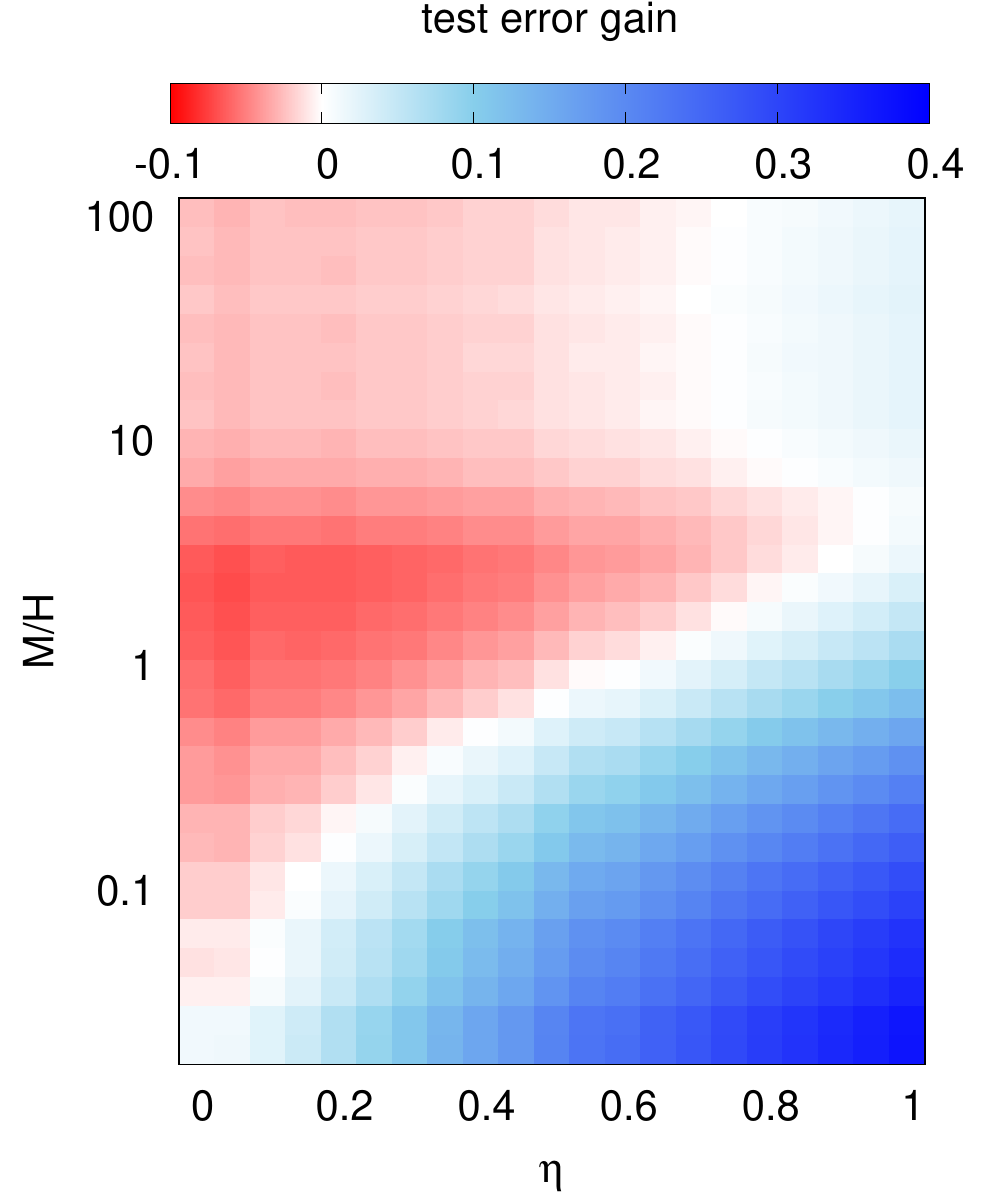}
        \caption{feature perturbation}    
    \end{subfigure}
    \begin{subfigure}[b]{0.33\textwidth}
        \includegraphics[width=\linewidth]{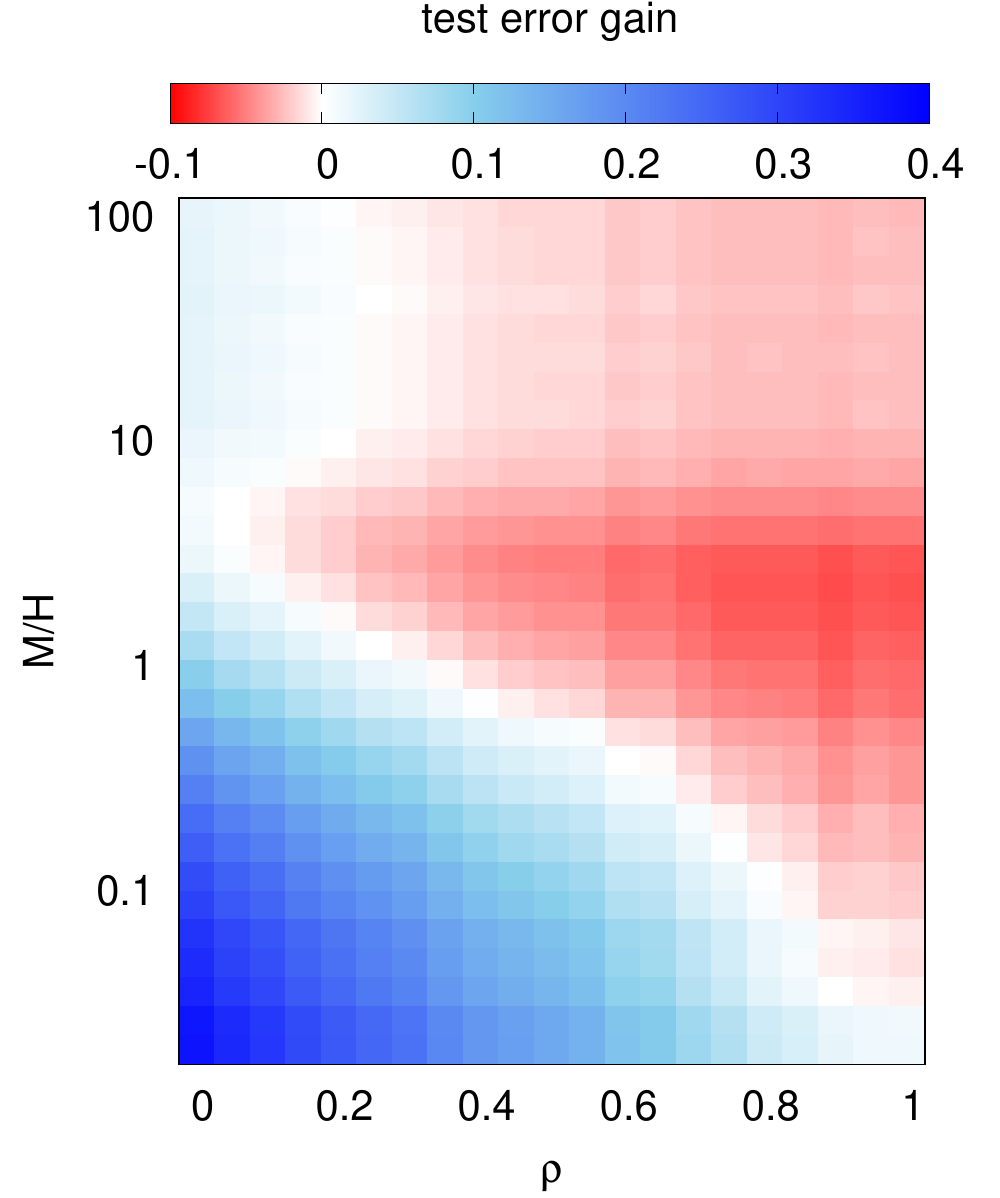}
        \caption{feature substitution}    
    \end{subfigure}
    
    \caption{\textbf{Transfer learning performance with correlated generative features}. The two phase diagrams characterize the learning performance in the CHMM, comparing the transferred feature model to a two-layer network with both layers trained on the target task (TF to 2L gain), as a function of: (a) the feature perturbation parameter $\eta$, (b) the feature substitution parameter $\rho$,  and the target dataset size $M$. The results for TF are obtained from the theoretical solution of the CHMM, while 2L is purely numerical ($10$ samples per point). Parameters: $L=200$, $q=1$, $D=1000$ and $H=500$.}
    \label{fig:feature_pertubations_phase_diagram}
\end{figure}

Fig~\ref{fig:feature_pertubations_phase_diagram} shows two phase diagrams, comparing TF with 2L as a function of: (a) the feature perturbation parameter $\eta$; (b) the feature substitution parameter $\rho$. High values of $\eta$ (low values of $\rho$) indicate strongly correlated tasks. Apart from the fact that the diagram associated to $\rho$ is mirror image compared to the others, it is evident that the obtained phase diagrams are non just qualitatively equivalent, but also quantitatively similar to the phase diagram for the parameter $q$. At low numbers of samples and high levels of feature correlation TF largely outperforms 2L. We also find again a region where TF overfits due to the closeness to the separability threshold. More generally, 2L becomes the better algorithm when the size of the target dataset is large enough.

The striking similarity of Fig.~\ref{fig:teacher_pertubations_phase_diagram}a and these phase diagrams might suggest some type of universality in the effect of dataset correlation (of any type) on the quality of the transferred features.

\paragraph*{Impact of the number of learned features.}
In this paragraph, we look at the effect of varying the the width of the two-layer neural network, $H$. Note that $H$ also corresponds to the number of learned features in a 2-layer network. 
In the plots, we rescale $H$ by the number of generative features (kept fixed to $L=200$) to obtain a quantity that remains $\mathcal{O}(1)$ even in the high-dimensional setting of the replica computation. 

\begin{figure}[H]
    \centering
    \begin{subfigure}[b]{0.33\textwidth}
        \includegraphics[width=\linewidth]{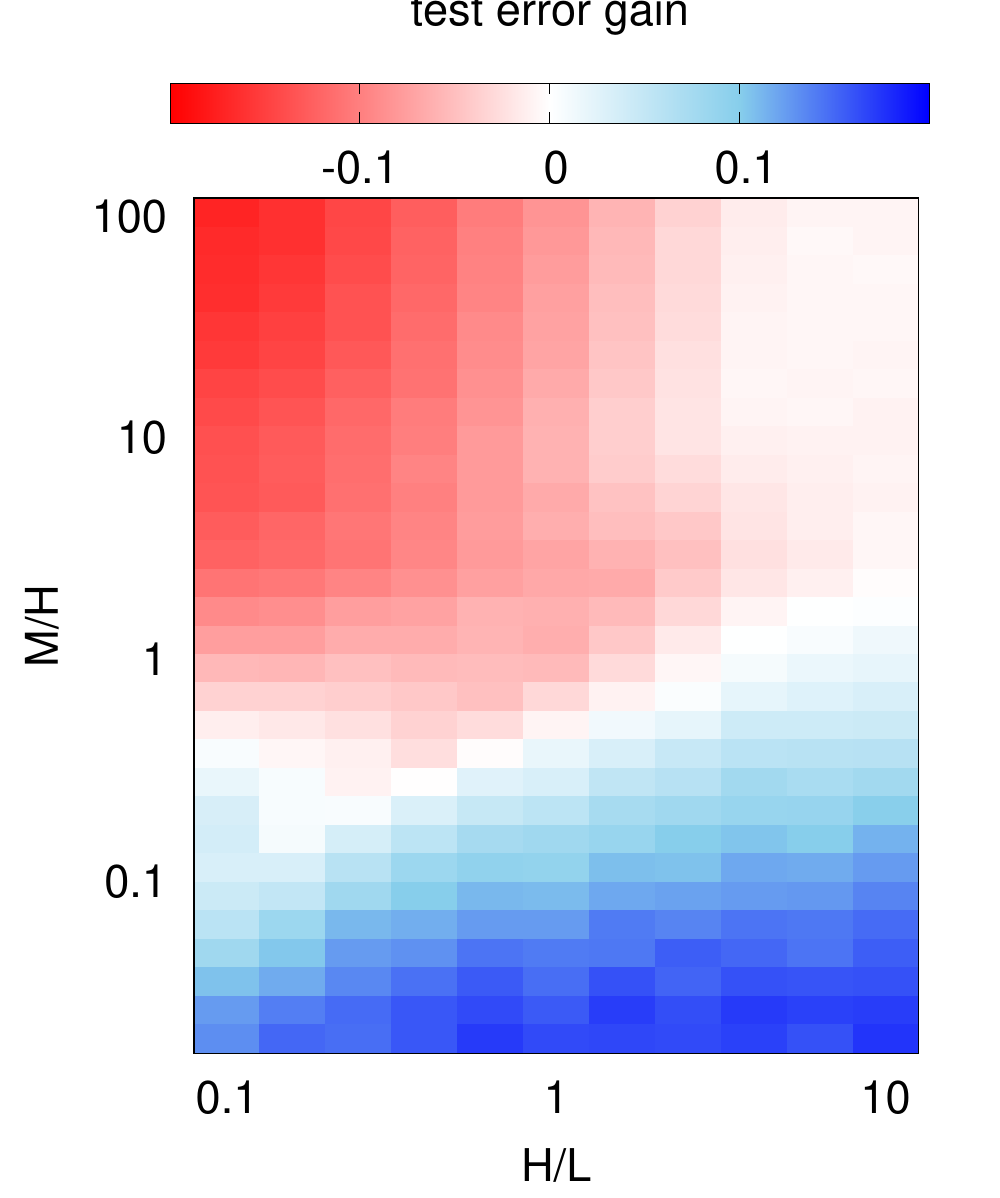}
        \caption{TF to 2L gain}    
    \end{subfigure}
    \begin{subfigure}[b]{0.33\textwidth}
        \includegraphics[width=\linewidth]{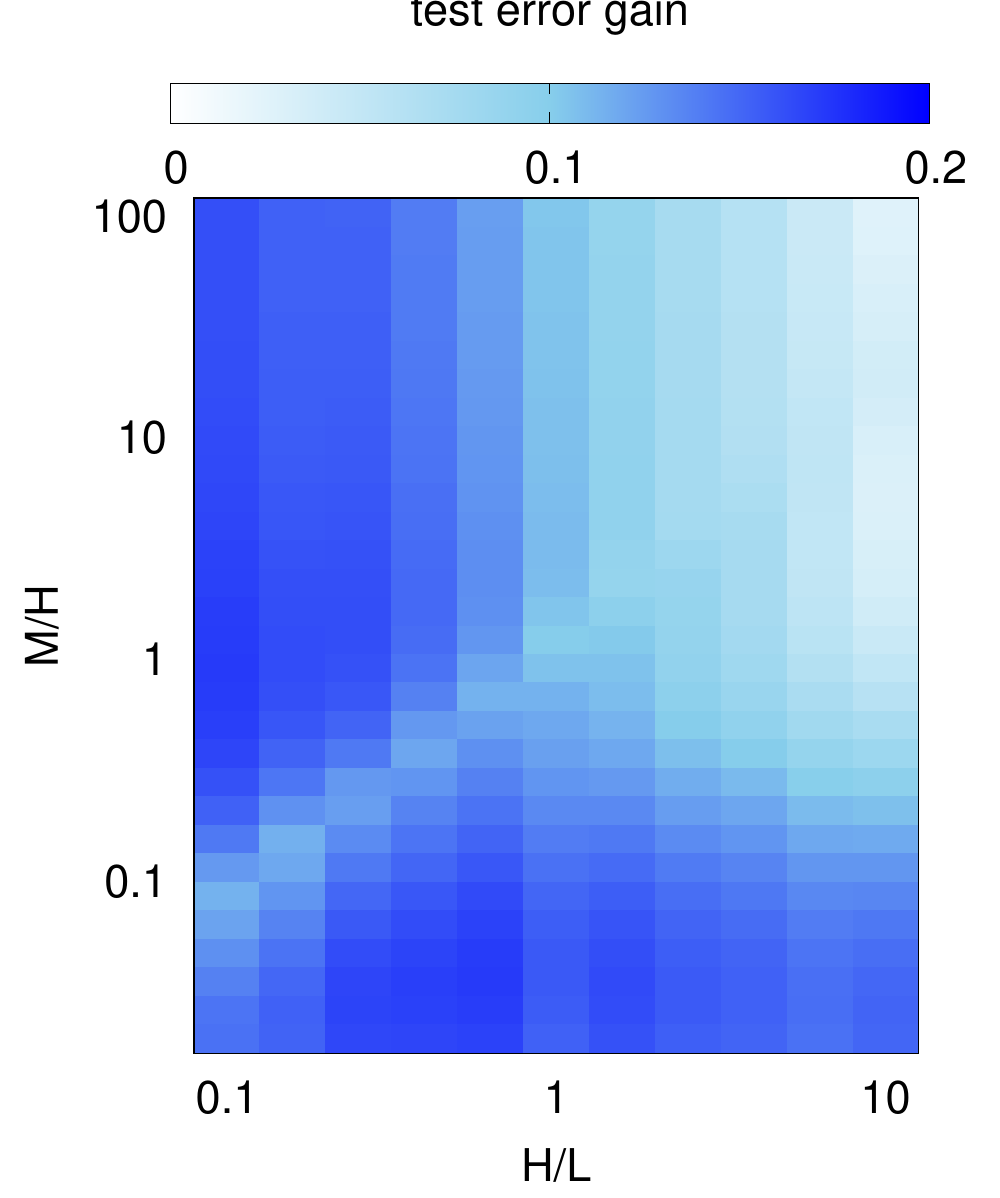}
        \caption{TF to RF gain}    
    \end{subfigure}
    
    \caption{\textbf{Impact of the hidden layer width on transfer learning performance}. The two phase diagrams compare the performance of the transferred feature model (TF) to (a) a two-layer network with both layers trained on the target task (2L), (b) a random feature model (RF), in the CHMM as a function of the number of hidden units $H$ and the number of samples in the target dataset $M$. The results for TF and RF are obtained from the theoretical solution of the CHMM, while 2L is purely numerical ($10$ samples per point). Parameters: $L=200$, $(\eta, \rho, q)=(1, 0.2, 0.8)$, and $D=1000$.}
    \label{fig:H_phase_diagram}
\end{figure}

Fig~\ref{fig:H_phase_diagram} shows two phase diagrams, comparing (a) TF with 2L and (b) TF with a RF, as a function of $H$ and of the number of samples in the target task. Source and target tasks are linked through a fixed transformation with parameters $(\eta, \rho, q)=(1, 0.2, 0.8)$. In both diagrams we can clearly see the diagonal line corresponding to the separability threshold,  which shifts to higher values as $H$ (i.e., the number of parameters in the second layer) is varied. TF is found to perform better in the low number of samples regime, as in all other experimental settings. An interesting phenomenon appears in panel (b), where the behavior of RF seems to change once the number of learned features becomes larger than the number of generative features $H>L$. From this point on, the improvement obtained by TF over RF becomes less pronounced if the number of samples is sufficient. Both diagrams also seem to show that at very large $H$, and starting from intermediate values of the number of samples, the differences between TF, 2L and RF seem to narrow. This type of behavior is expected, since by growing the width of the hidden layer one eventually approaches the kernel regime \cite{spigler2019asymptotic}.

\paragraph*{Impact of sample size of source dataset.}
Finally, we vary a parameter that was kept fixed throughout the above presented simulations, the number of samples in the source task. Of course, in practice it does not make too much sense to use a small dataset as a source task in a transfer learning procedure. The interest in this experiment is more theoretical, as it can be used to understand when the learning model is able to start extracting features that could be helpful in a different task. 

\begin{figure}[]
    \centering
    \begin{subfigure}[b]{0.33\textwidth}
        \includegraphics[width=\linewidth]{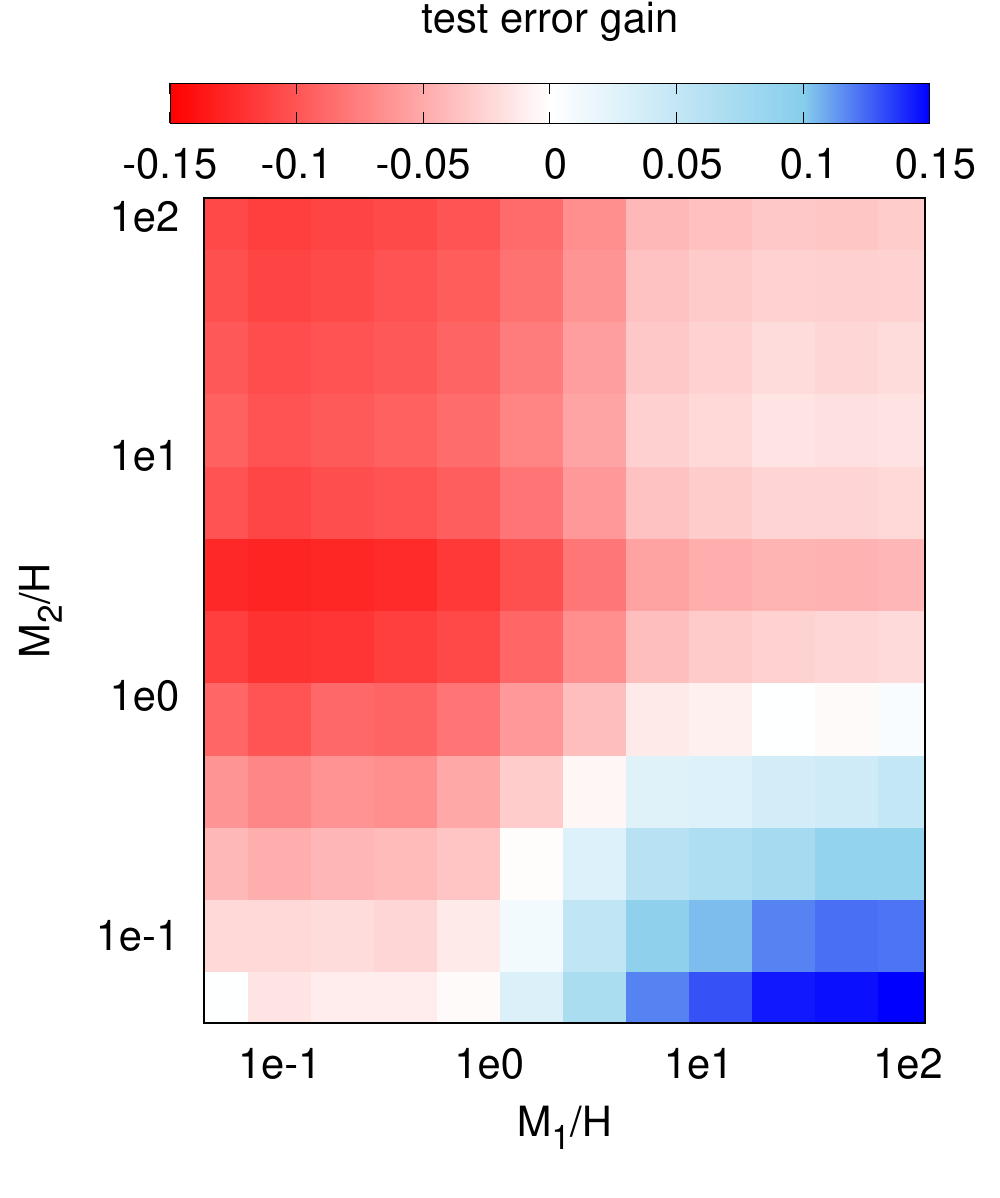}
        \caption{TF to 2L gain}    
    \end{subfigure}
    \begin{subfigure}[b]{0.33\textwidth}
        \includegraphics[width=\linewidth]{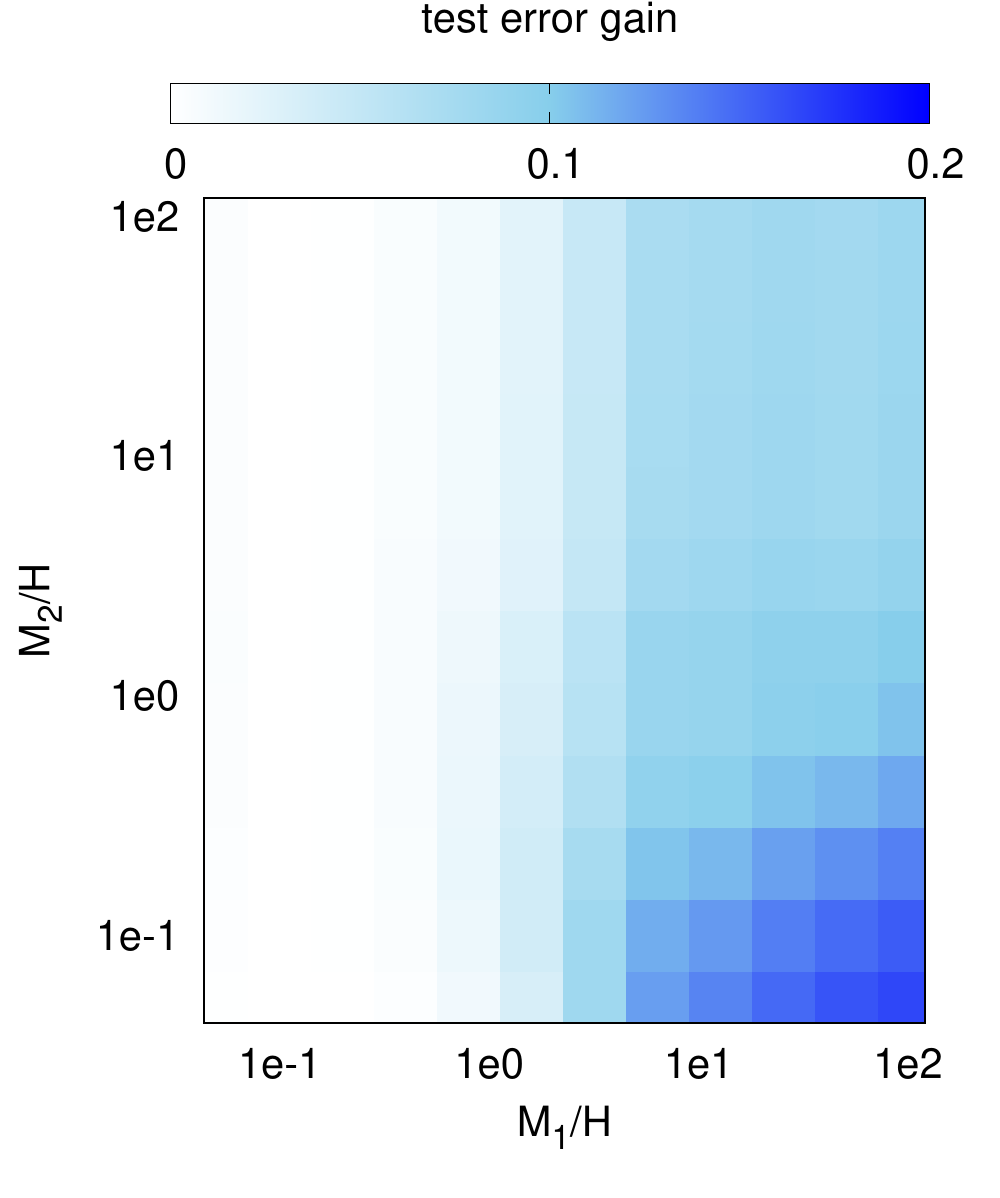}
        \caption{TF to RF gain}    
    \end{subfigure}
    
    \caption{\textbf{Impact of the source dataset size on transfer learning performance}. The two phase diagrams compare the performance on the CHMM of the transferred feature model (TF) to (a) a two-layer network with both layers trained on the target task (2L), (b) a random feature model (RF), as a function of the number of samples in the source dataset $M_1$, and the number of samples in the target dataset $M_2$. The results for TF and RF are obtained from the theoretical solution of the CHMM, while 2L is purely numerical ($10$ samples per point). Parameters: $L=200$, $(\eta, \rho, q)=(1, 0.2, 0.8)$, $D=1000$ and $H=500$.}
    \label{fig:M_phase_diagram}
\end{figure}

Fig~\ref{fig:M_phase_diagram} offers a comparison of (a) TF with 2L and (b) TF with a RF, as a function of source and target dataset sizes, $M_1$ and $M_2$. The resulting phase diagram shows that the two-layer network trained on the source dataset is unable to extract good features from the data below a critical value of the number of samples $M_1$. In this regime the performance of TF in the target task is indistinguishable from RF, confirming the key role played by the learned features (and not just initialization and scaling \cite{dhifallah2021phase}) in the success of TF. As expected, when the number of samples in the second task becomes larger, we reconnect with the typical scenario already recorded in the previous experiments.  

\section{Numerical experiments on multi-label learning}\label{app:multiclass} 

In the proposed CHMM model we focus on the case of binary classification. One of the main reasons for this choice is that the asymptotic equations derived through our analysis can in this case be solved in reasonable time, while adding extra dimensions in the output space would likely make their repeated solution too computationally demanding. Although a majority of the related theoretical works is similarly limited to the simplified binary case, intuitively the effect of cross-class correlation and feature similarity among multiple classes could have a large impact on transfer learning performance. On the other hand, it has been recently observed that multi-label classification shows similar phenomenology to the case of binary classification \cite{loureiro2021learning}, supporting the assumption that the study of the simpler case may also allow for a deeper understanding of the general case. 

In this section we test the robustness of our results in the multi-label classification settings making a simple extension of the CHMM model. This test is performed numerically and does not have the purpose of extending the theory presented in the main text (although that may be possible, in principle, by extending our results in the spirit of \cite{loureiro2021learning}). Note that, because of the pure numerical nature of these experiments, the computational cost of the associated simulations was greatly increased with respect to the experiments presented in the main text, and required an implementation on a computational cluster.

The multi-label CHMM model can be again represented using the cartoon shown in Fig.~\ref{fig:model_cartoon} with the difference that $y$ and $\hat y$ are $K$-dimensional arrays, with $K$ the number of classes. The correct label is given by the highest coordinate of $y$, while the output of the student is re-normalized using a softmax function. Perturbations between source and target tasks can be defined likewise.

\begin{figure}[h]
    \centering
    \begin{subfigure}[b]{0.325\textwidth}
        \includegraphics[width=\linewidth]{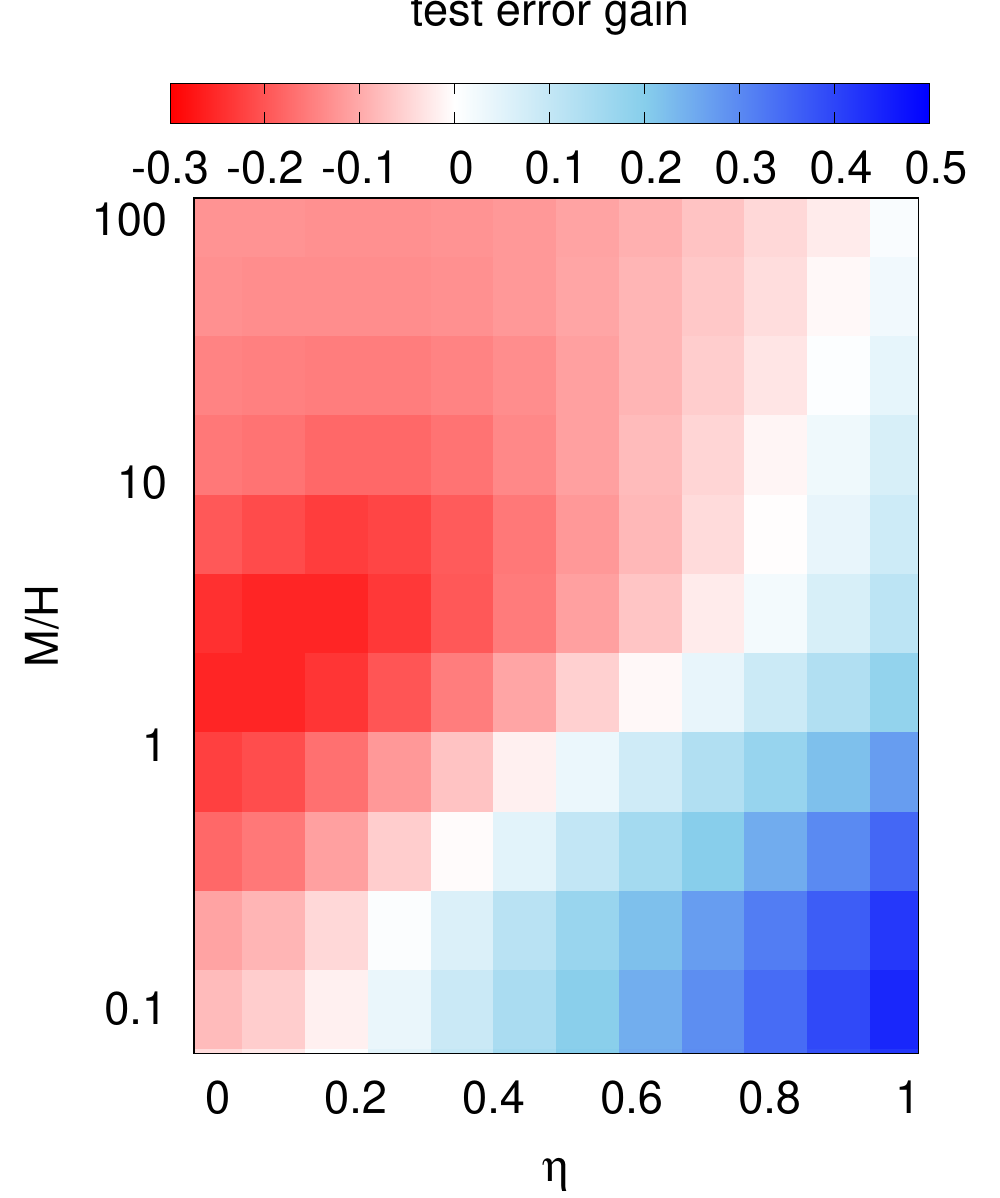}
        \caption{TF to 2L gain}
    \end{subfigure}
    \begin{subfigure}[b]{0.325\textwidth}
        \includegraphics[width=\linewidth]{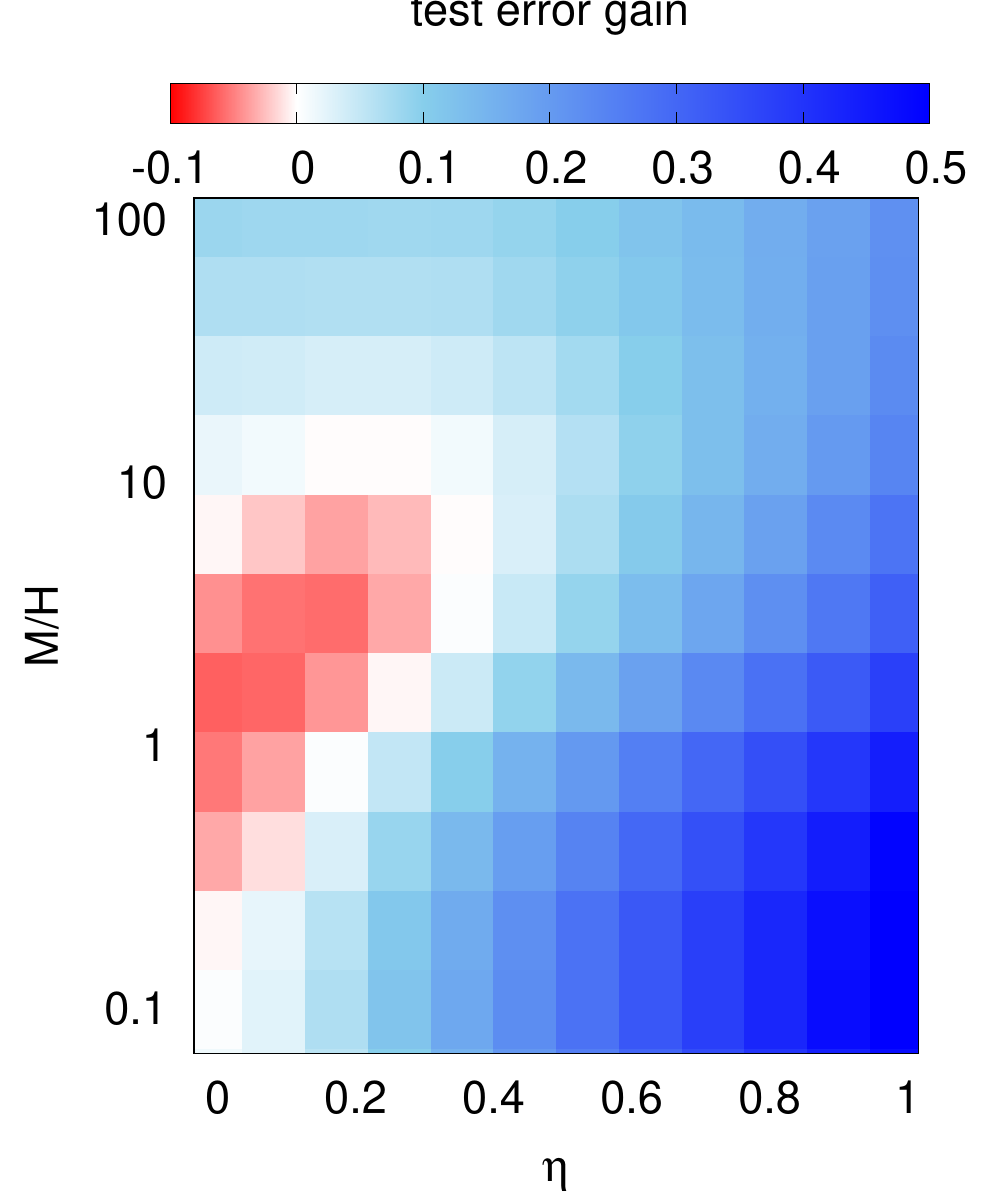}
        \caption{TF to RF gain}
    \end{subfigure}
    \begin{subfigure}[b]{0.325\textwidth}
        \includegraphics[width=\linewidth]{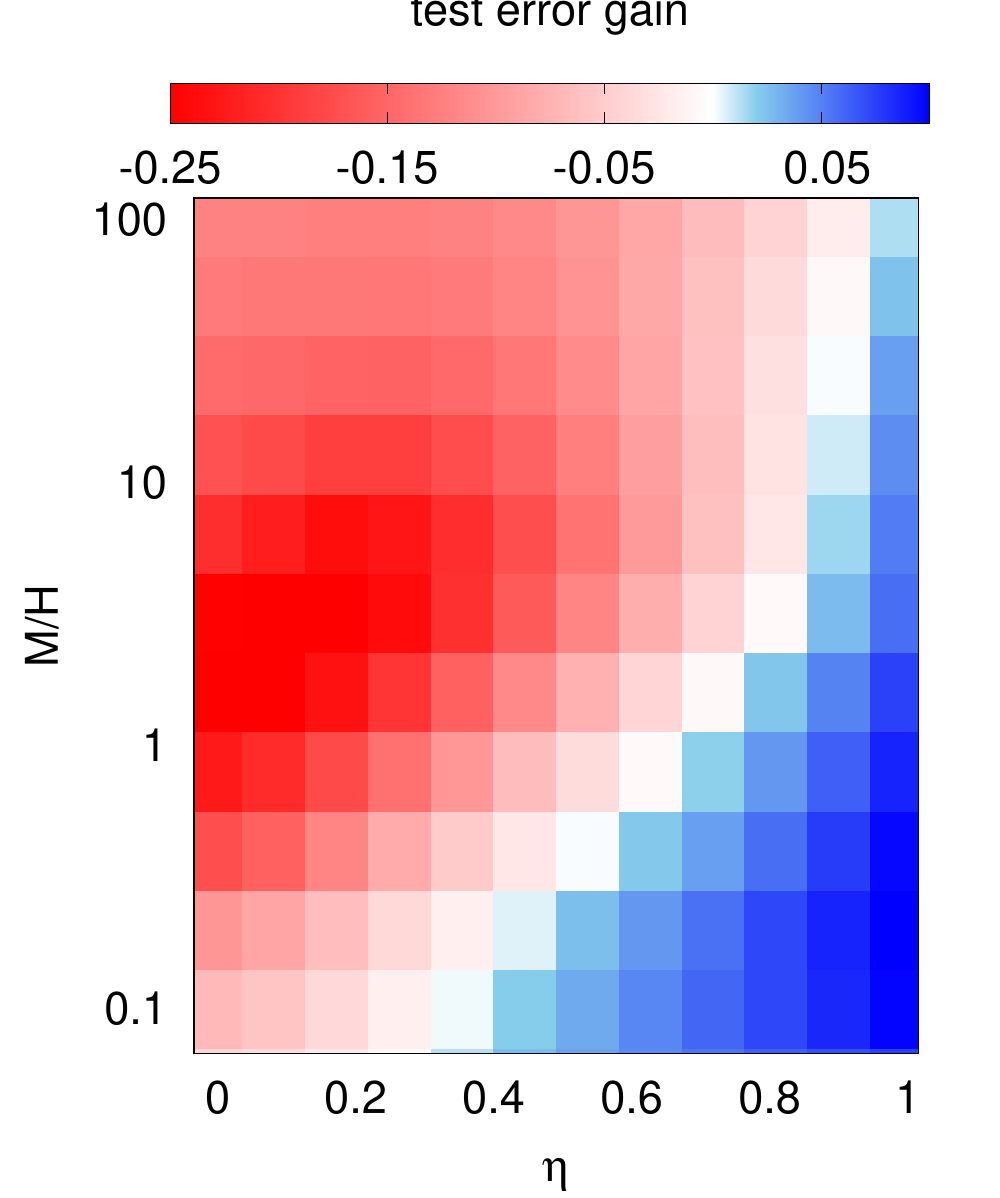}  
        \caption{TF to ft-TF gain}
    \end{subfigure}
    \caption{\textbf{Effect of the dataset correlation on transfer learning performance in multi-label classification}. The three phase diagrams characterize the learning performance in the CHMM as reported in Fig.~\ref{fig:teacher_pertubations_phase_diagram} as $\eta$ is varied. The gain is obtained after averaging over 10 realizations. Parameters: $L=200$, $(q, \rho)=(1,0)$, $D=1000$, $H=500$ and $K=4$.}
    \label{fig:teacher_pertubations_phase_diagram_multiclass}
\end{figure}

Fig.~\ref{fig:teacher_pertubations_phase_diagram_multiclass} is the multi-label counterpart of Fig.~\ref{fig:feature_pertubations_phase_diagram}a (and Fig.~\ref{fig:teacher_pertubations_phase_diagram}, given the quantitative and qualitative similarity between the phase diagrams for the parameters $\eta$ and $q$). The main result we report is that, despite some quantitative disagreement, the salient features of the diagrams are still there, as expected. The simulations were purposely evaluated with minimal changes with respect to the binary case. The comparison between TF and RF shows again negative transfer for low $\eta$ and low sample complexity in the target, Fig.~\ref{fig:teacher_pertubations_phase_diagram_multiclass}b. The different shape of the region may be due to poor feature extraction in the numerical optimization, indicating that a larger number of learning steps may be needed. The phase diagrams on TF vs 2L (Fig.~\ref{fig:teacher_pertubations_phase_diagram_multiclass}a) and TF vs ft-TF (Fig.~\ref{fig:teacher_pertubations_phase_diagram_multiclass}c) show the largest impact of transfer learning when the number of samples in the target dataset is low and when the similarity between the two sets is large enough. 

\begin{figure}[h]
    \centering
    \begin{subfigure}[b]{0.38\textwidth}
        \includegraphics[width=\linewidth]{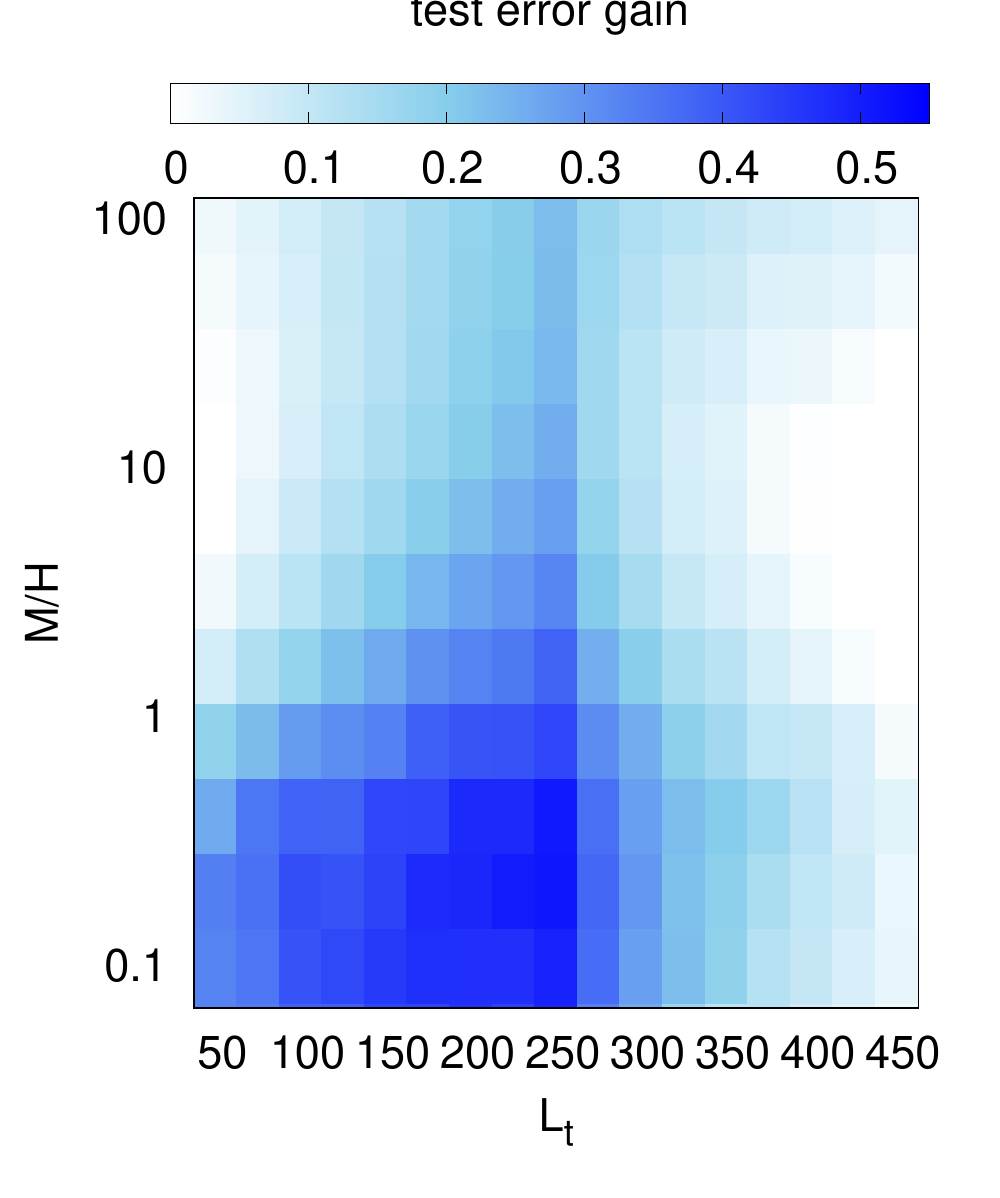}
        \caption{TF to RF gain}
    \end{subfigure}
    \begin{subfigure}[b]{0.58\textwidth}
        \includegraphics[width=\linewidth]{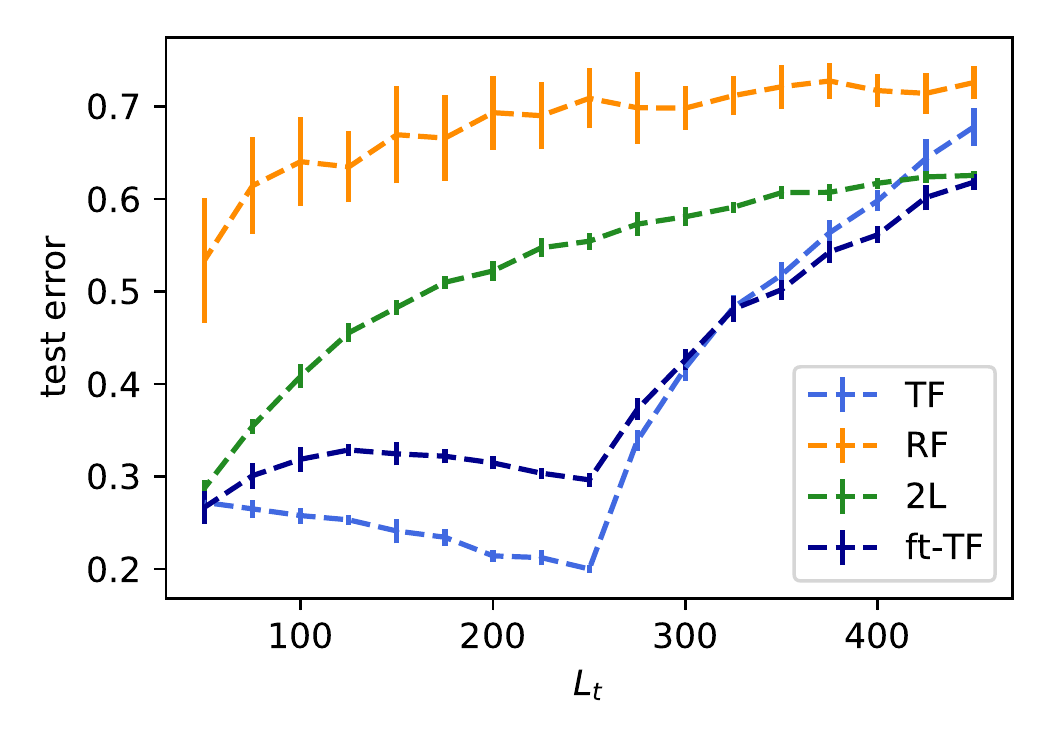}  
        \caption{horizontal slice of the phase diagram}
    \end{subfigure}
    \caption{\textbf{Effect of a latent dimension asymmetry in multi-label classification}. The three phase diagrams characterize the learning performance in the CHMM as reported in Fig.~\ref{fig:asymmetric_phase_diagram}. The plot in the multi-label setting shows qualitative agreement to the binary case.
    Parameters: $L_s+L_t=500$, $(q, \rho, \eta)=(1,0,1)$, $D=1000$, $H=500$ and $K=4$.
    }
    \label{fig:asymmetric_phase_diagram_multiclass}
\end{figure}

Let us consider the case of asymmetric transfer, where the target and source hidden dimensions ($L_s$ and $L_t$ respectively) differ, extending the analysis of Sec.~\ref{sec:asymmetric} to the multi-label cases. Fig.~\ref{fig:asymmetric_phase_diagram_multiclass} shows the same features of Fig.~\ref{fig:asymmetric_phase_diagram}, the most beneficial case occurs for low sample complexity and transferring from high dimension to low dimension ($L_s>L_t$). This is consistent with the literature and correctly captured by the model.

Finally, we again remark that the key advantage of the standard binary CHMM is that, because of its simplicity and the existence of universality, it allows for a fast and exact characterization of the same phenomena in a short time, without the need of computationally expensive simulations. 

\section{More experiments on real data}\label{app:exp_real_data}

In the following paragraphs, we provide some additional transfer learning experiments on real data. The goal of this section is to show that, despite different types of relationship between source and target tasks and different degrees of relatedness, the emerging qualitative behavior is similar to that presented in the main text and reproduced by the CHMM. 
In paragraph 1 of section \ref{sec:result}, we have considered an experiment corresponding to a \emph{feature substitution} transformation in the context of the CHMM. In the following, we will describe two experimental designs that instead correspond to \emph{feature perturbation} and \emph{teacher perturbation} transformations (see section \ref{sec:problem} for more details). Moreover, we will demonstrate a case of ``orthogonal'' tasks, where the feature transfer is completely ineffective.

\paragraph*{Feature perturbation.} We consider an experiment where the transfer is between the MNIST dataset and a perturbed version of MNIST, obtained by applying a data-augmentation transformation to each image. In particular, we construct the source dataset, $\mathcal{D}_s$, by altering the MNIST dataset through an edge enhancer in the \emph{imgaug} library for data augmentation (for more details see sec.\ref{app:simulation_details}). Instead, we use the original MNIST as the target dataset, $\mathcal{D}_t$. In both cases, we label the images according to even-odd digits. In this way, the target task can be seen as a perturbed version of the source task.    

\begin{figure}[h]
    \centering
    \begin{subfigure}[m]{0.49\textwidth}
        \includegraphics[width=\linewidth]{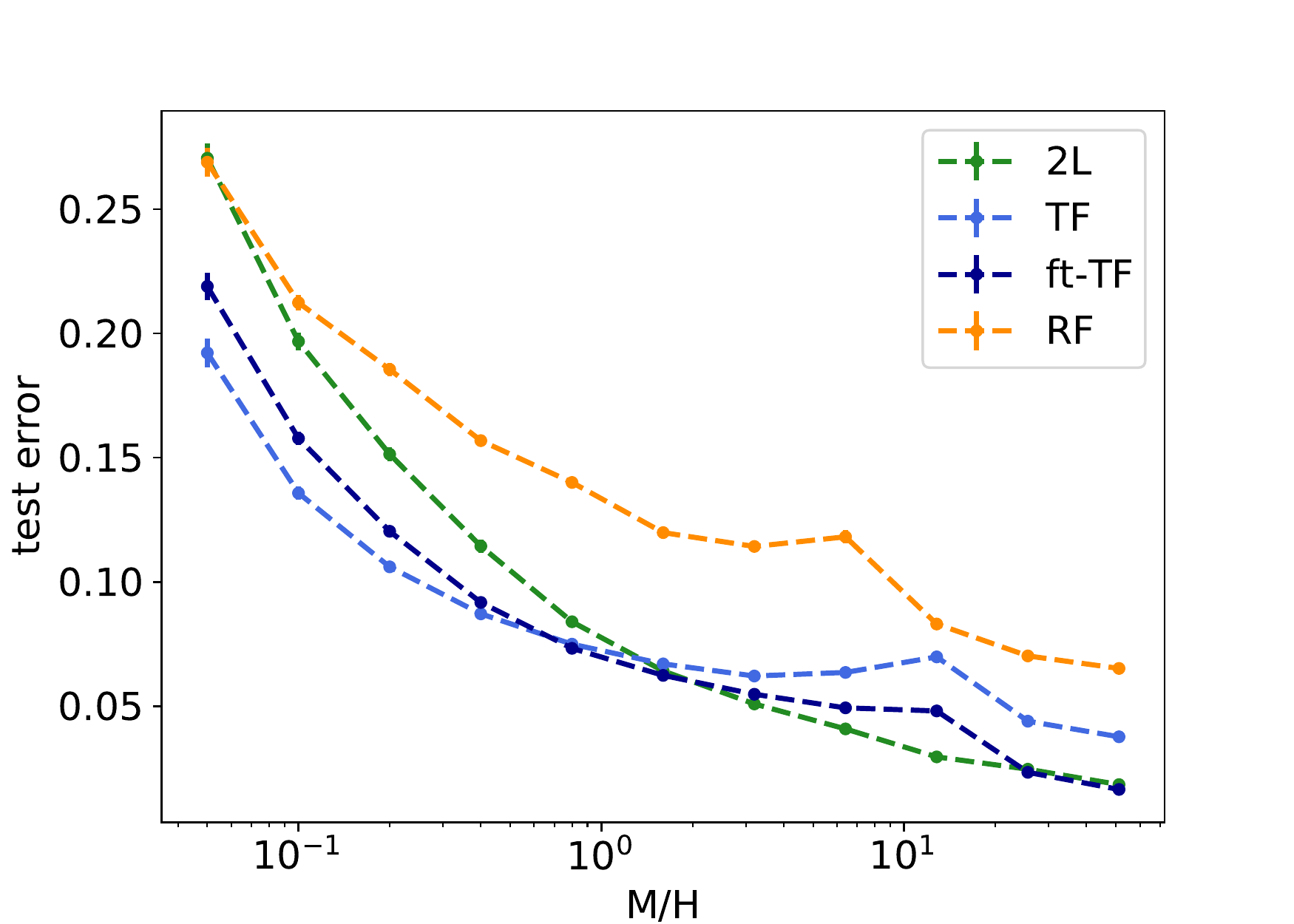}
        \caption{feature perturbation}    
    \end{subfigure}
    \caption{\textbf{Feature perturbation}: test error as a function of the number of samples per hidden unit. The source dataset is the augmented even-odd MNIST, while the target dataset is the original even-odd MNIST. The employed augmentation is an edge-enhancer transformation. The parameters of the networks are set to: $D = 784$, $H = 500$ and $\lambda = 1e-8$. The size of the source task is $M_s = 60000$. All the experiments have been averaged over $50$, $20$ and $10$ samples for low, intermediate and high numbers of samples respectively.}
    \label{fig:feature_perturbation}
\end{figure}

Fig.~\ref{fig:feature_perturbation} shows the outcome of this experiment. A first thing to notice is that TF always outperforms RF for all numbers of samples in the range we considered. Moreover, both models show a double descent behavior, with the peak occurring at the linear separability threshold (see sec.~\ref{sec:separability_threshold} for more details). As seen in the main text, TF reveals to be more effective than 2L at small numbers of samples, with a gain of about $5\%$ in test error scores. This gap closes above $M/H = 1$, where the 2L starts performing better than both TF and ft-TF. Concerning ft-TF, we can instead recognize two different regimes. At small number of samples, ft-TF performs better than 2L but worse than TF, showing a larger over-fitting effect compared to Fig.~\ref{fig:teacher_pert_real_vs_synthetic}. With a higher number of samples, ft-TF joins 2L, thus outperforming TF, except in the small region close to the separability threshold.

\paragraph*{Teacher perturbation and orthogonal tasks.} Additionally, we consider the following experiment on real data. As a source dataset, $\mathcal{D}_s$, we use the notMNIST dataset \cite{notmnist} grouped into even-odd classes. In the target task, $\mathcal{D}_t$, we instead consider notMNIST but labeled differently, based on whether the class labels are smaller or larger than $5$. In this way, the only difference in the two tasks is in the employed labeling rule. 

\begin{figure}[]
    \centering
    \begin{subfigure}[m]{0.49\textwidth}
        \includegraphics[width=\linewidth]{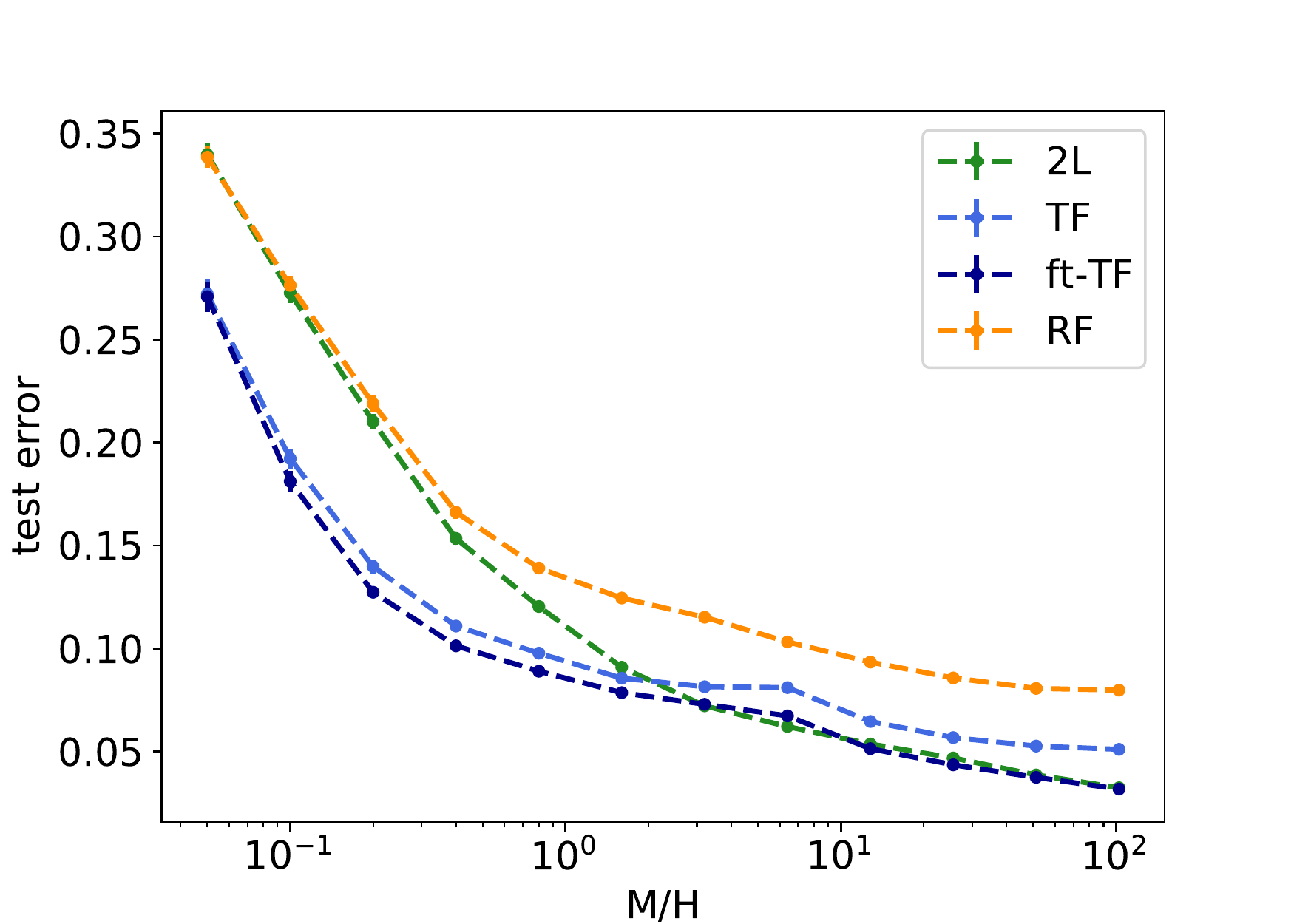}
        \caption{teacher perturbation}
    \end{subfigure}
    \begin{subfigure}[m]{0.49\textwidth}
        \includegraphics[width=\linewidth]{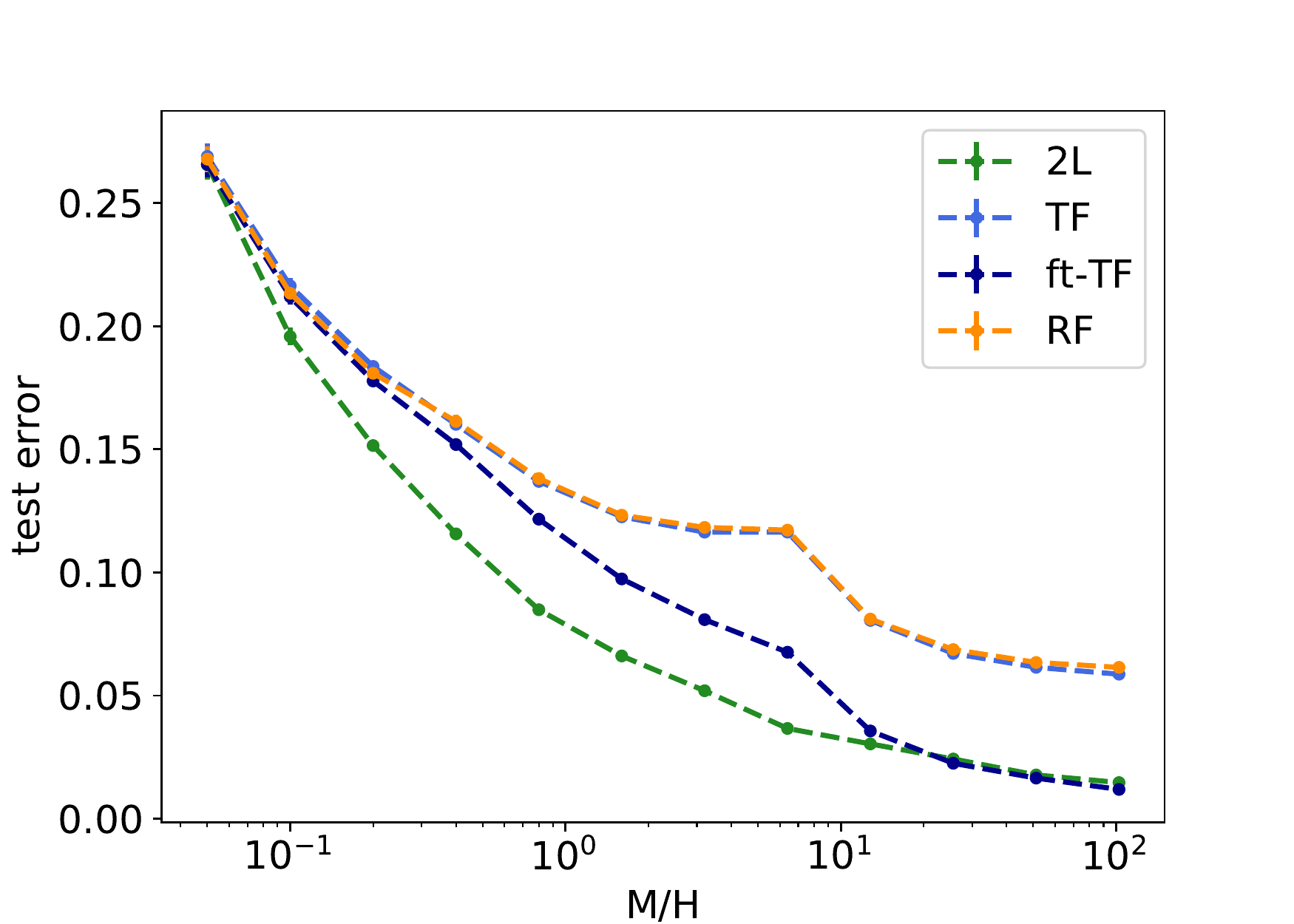}
        \caption{orthogonal teachers}
    \end{subfigure}
    
    \caption{\textbf{Perturbed teacher and orthogonal tasks:} Generalization error as function of the number of samples per hidden unit in the target task. (a) teacher perturbation. (b) orthogonal tasks. In the left panel, the source dataset is the even-odd notMNIST, while the target dataset is notMNIST grouped in digits smaller-greater than 5. The size of the source task is $M_s = 200000$. In the right panel, the source task is MNIST grouped according to image luminosity and the target task is even-odd MNIST. The size of the source task is $M_s = 51200$. Note that in panel (b), the light blue and the orange curves are overlapping. The parameters of the networks are set to: $D = 784$, $H=500$ and $\lambda=1e-8$. The simulations are averaged over 50, 20, and 10 samples at small, intermediate and high numbers of samples respectively.}
    \label{fig:teacher_perturb_ort}
\end{figure}

Fig.~\ref{fig:teacher_perturb_ort}(a) displays the outcome of this experiment. The observed phenomenology is identical to that already described for the previous experiments. Note that, even though the two tasks share the same input images, observing a benefit when transferring the feature map is not trivial. TF is here found to be effective because the learned features carry information about the digit represented in each image, which induces a useful representation regardless of the grouping of the digits.

It is in fact possible to construct effectively ``orthogonal'' tasks even if the set of inputs is the same. In the final experiment, we use even-odd MNIST as the target task, $\mathcal{D}_t$, but we consider a source task where the label assigned to each MNIST image is only dependent on its luminosity. In particular, we assign all the images with average brightness less than $20$ or in the interval $(35,59)$ to one group and the remaining ones to the other group. Thus, the resulting source task has nothing to do with digit recognition. 

Fig.~\ref{fig:teacher_perturb_ort}(b) shows the outcome of this last experiment. Contrary to the other cases, in this experiment we can clearly see no advantage in transferring the feature map from the source to the target task: TF not only does not improve over 2L, but it also overlaps with RF for any number of samples. Interestingly, we can here identify three different regimes for ft-TF: a first regime at small numbers of samples where ft-TF actually follows the trend of TF and RF, thus badly performing with respect to 2L; a second regime, at intermediate numbers of samples, where ft-TF actually starts improving its test scores with respect to TF and RF (around $10\%$) but it still does not have enough data to perform as well as 2L; finally, a third regime when the number of samples is high, where ft-TF equates the generalization performances of 2L.     

\section{Technical details on the numerical simulations}\label{app:simulation_details} 

In this section, we provide additional details on the numerical simulations concerning the experiments on both real and synthetic data. 

\begin{figure}[h]
    \centering
    \begin{subfigure}[t]{0.49\textwidth}
        \includegraphics[width=\linewidth]{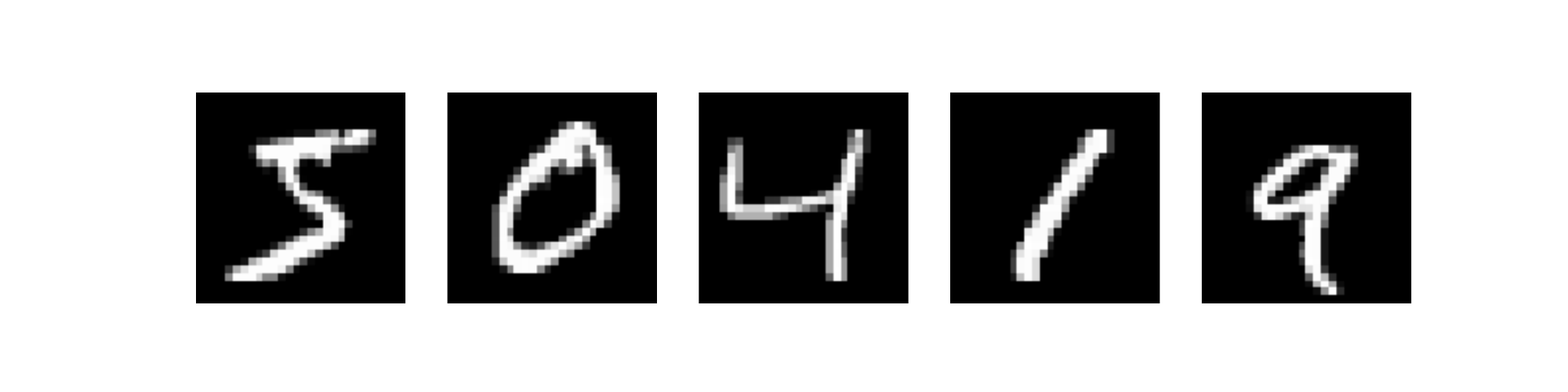}
        \caption{MNIST}
    \end{subfigure}
    \begin{subfigure}[t]{0.49\textwidth}
        \includegraphics[width=\linewidth]{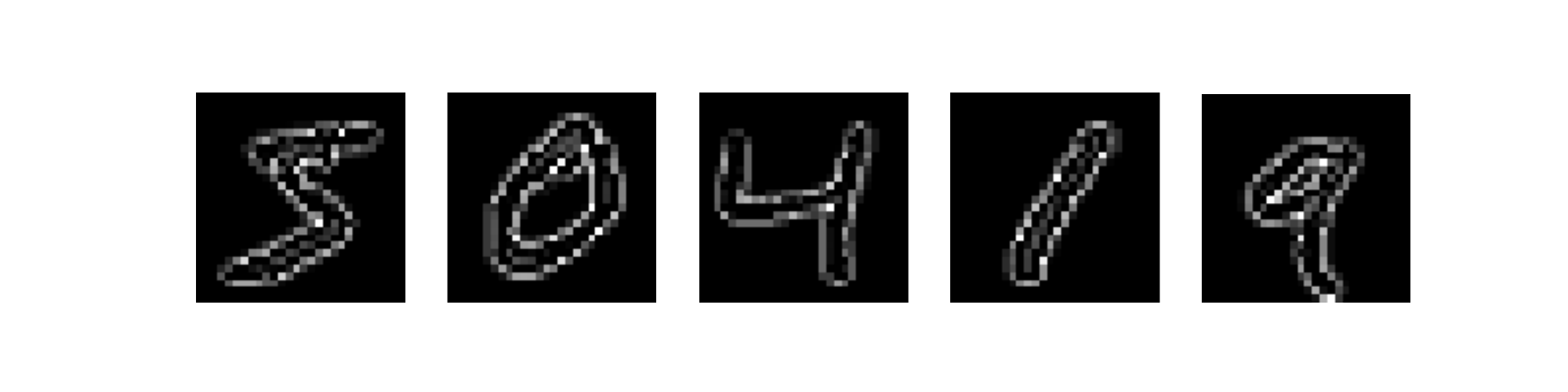}
        \caption{MNIST augmented}
    \end{subfigure}
    \begin{subfigure}[b]{0.49\textwidth}
        \includegraphics[width=\linewidth]{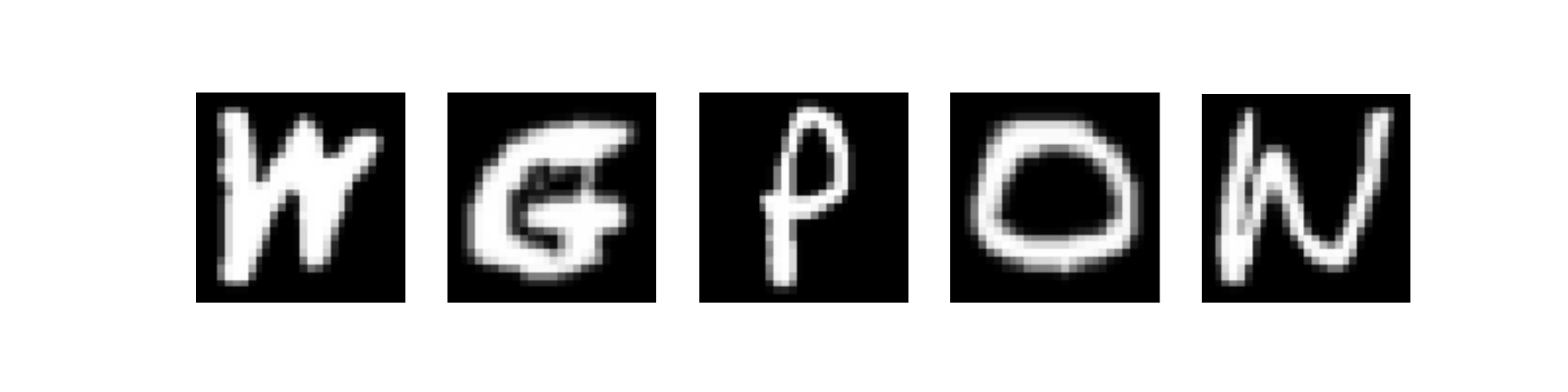}
        \caption{EMNIST-letters}
    \end{subfigure}
    \begin{subfigure}[b]{0.49\textwidth}
        \includegraphics[width=\linewidth]{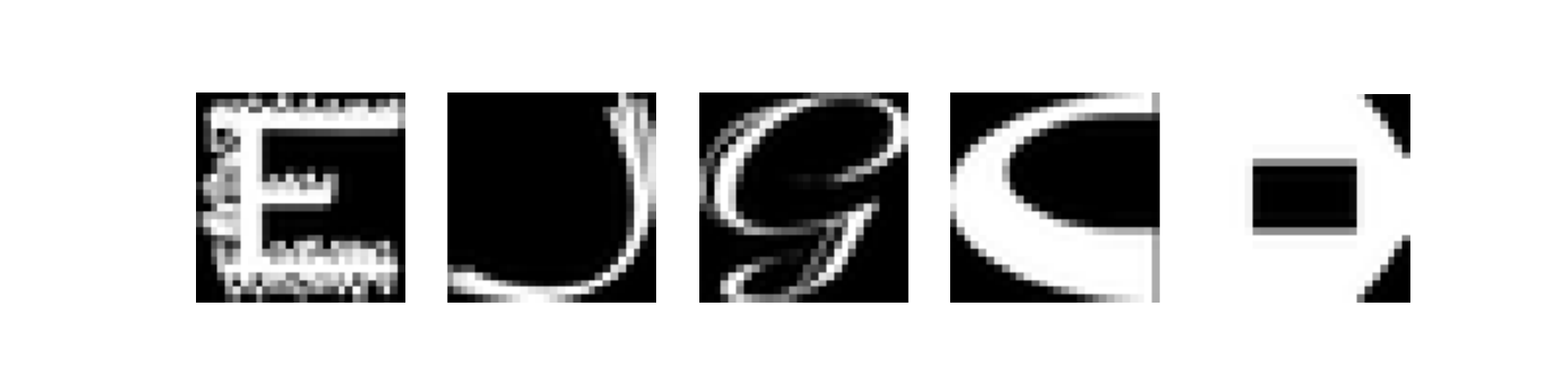}
        \caption{notMNIST}
    \end{subfigure}
    \caption{\textbf{Examples from the real datasets:} (a) examples of images in the MNIST dataset. (b) examples of images in the MNIST dataset perturbed to enhance digit contours with respect to the background. (c) examples of images in the EMNIST-letters dataset. (d) examples of images in the notMNIST dataset.}
    \label{fig:examples_of_datasets}
\end{figure}

\paragraph*{Datasets.} In the experiments with real data, we have used three different standard datasets: MNIST, EMNIST-letters and notMNIST. MNIST is a database of images ($28 \times 28$ pixels in range $0-255$ in the experiments) of handwritten digits with $10$ classes, containing $60000$ examples in the training set and $10000$ examples in the test set \cite{deng2012mnist}. EMNIST-letters is a dataset of images ($28 \times 28$ pixels in range $0-1$ in the experiments) of handwritten letters with $26$ classes, containing $124800$ examples in the training set and $20800$ examples in the test set \cite{cohen2017emnist}. Finally, notMnist is a dataset of images ($28 \times 28$ pixels in range $0-1$ in the experiments) of letter fonts from $A$ to $J$ with $10$ classes, containing $200000$ examples in the training set and $10000$ examples in the test set \cite{notmnist}. Concerning the experiment on feature perturbation of Fig.~\ref{fig:feature_perturbation}, we have used the \emph{EdgeDetect} function of the \emph{imgaug} library for data augmentation to enhance the contours of each digit with respect to the background \cite{imgaug}. In the experiment, we set the parameter \emph{alpha} to $(0.5, 0.7)$. An example of the three datasets is provided in Fig.~\ref{fig:examples_of_datasets}.  

\paragraph*{Hyper-parameter settings.}
In the source task, we consider a two-layer neural network with ReLU activation function and single sigmoidal output unit. To train the two-layer network on the source, we implement mini-batch gradient descent, using the end-to-end open source platform for Machine Learning Tensorflow 2.4 \cite{tensorflow2015-whitepaper}. In particular, we use the Adam optimizer with default Tensorflow hyper-parameters and the binary cross-entropy loss. We then apply $L_2$ regularization on the last layer and early-stopping with default Tensorflow hyper-parameters as regularizers. The training is immediately stopped when an increase in the test loss is recorded (the \emph{patience} parameter in early stopping is set to zero). We set the maximum number of epochs to $200$, the learning rate to $1e-3$ and the batch size to $50$. 

In the target task, we train TF and RF via the scikit-learn open-source python library for Machine Learning \cite{scikit-learn}. In particular, we use the module $linear\_model$.$LogisticRegression$ which implements logistic regression with $L_2$ regularization when the \emph{penalty} parameter is set to ``l2''. Note that, non-zero L2 regularization ensures that the optimization process is bounded even below the separability threshold for both TF and RF. The training stops either because a maximum number of iterations ($\text{max iters} = 10^4$) has been reached, or because the maximum component of the gradient is found to be below a certain threshold ($\text{tol} = 1e-7$). To train 2L, we have instead employed Tensorflow 2.4 once again, with Adam optimizer and cross-entropy loss with $L_2$-regularization on the last layer. In this case, we set the maximum number of epochs to $200$, the learning rate to $0.1$ and the batch-size to $1000$. The training stops when the maximum number of epochs is reached. The choice of the learning hyper-parameters is made to ensure the two-layer network to always reach zero-training error on the target task. Finally, to train ft-TF (from the TF initialization), we use the Adam optimizer and the binary cross-entropy loss. Since we expect the pre-trained weights on the source task to be already good enough to ensure good generalization performances, we set the learning rate to $0.01$ not to alter them too much or too quickly. We then keep the total number of epochs equal to $200$ and the batch-size equal to $1000$).

\end{document}